\pdfoutput = 1

\documentclass[11pt]{article}
\usepackage{fullpage}
\usepackage{comment}
\usepackage[utf8]{inputenc} 
\usepackage[T1]{fontenc}    
\usepackage{hyperref}       
\usepackage{url}            
\usepackage{booktabs}       
\usepackage{amsfonts}       
\usepackage{nicefrac}       
\usepackage{microtype}      
\usepackage{xcolor}         
\usepackage[pdftex]{graphicx}
\usepackage{subcaption}
\usepackage{caption}

\usepackage{amsfonts}       
\usepackage{amssymb}
\usepackage{amsmath}
\usepackage{amsthm}
\usepackage{mathtools}
\usepackage{multicol}
\usepackage{multirow}
\usepackage{amsthm}
\newtheorem{lem}{Lemma}
\newtheorem{prop}{Proposition}
\theoremstyle{definition}

\newtheorem{defn}{Definition}
\newtheorem{rem}{Remark}
\usepackage{color}

\usepackage{array}
\usepackage{algorithm}
\usepackage{algorithmic}

\def\vec{{\mathrm{vec}}}

\newcommand{\trace}{\mathrm{tr}}

\usepackage{soul}
\newcounter{guocomm}

\newcommand{\guorm}[1]{\ignorespaces}

\newcommand{\EScomment}[1]{{\color{blue} #1}}

\usepackage[utf8]{inputenc}

\date{}

\title{\textbf{How Curvature Enhance the Adaptation Power of Framelet GCNs}}

\author{Dai Shi\footnote{Western Sydney Univeristy,
\texttt{20195423@student.westernsydney.edu.au}, 
\texttt{y.guo@westernsydney.edu.au}}, \, Yi Guo, \,
Zhiqi Shao\footnote{University of Sydney,
\texttt{zsha2911@uni.sydney.edu.au}, \texttt{junbin.gao@sydney.edu.au}},\, Junbin Gao}

\begin{document}
\maketitle
\begin{abstract}
Graph neural network (GNN) has been demonstrated powerful in modeling graph-structured data. However, despite many successful cases of applying GNNs to various graph classification and prediction tasks, whether the graph geometrical information has been fully exploited to enhance the learning performance of GNNs is not yet well understood. This paper introduces a new approach to enhance GNN by discrete graph Ricci curvature. Specifically, the graph Ricci curvature defined on the edges of a graph measures how difficult the information transits on one edge from one node to another based on their neighborhoods. Motivated by the geometric analogy of Ricci curvature in the graph setting, we prove that by inserting the curvature information with different carefully designed transformation function $\zeta$, several known computational issues in GNN such as over-smoothing can be alleviated in our proposed model. Furthermore, we verified that edges with very positive Ricci curvature (i.e., $\kappa_{i,j} \approx 1$) are preferred to be dropped to enhance model's adaption to heterophily graph and one curvature based graph edge drop algorithm is proposed. Comprehensive experiments show that our curvature-based GNN model outperforms the state-of-the-art baselines in both homophily and heterophily graph datasets, indicating the effectiveness of involving graph geometric information in GNNs.
\end{abstract}








\section{Introduction}
Graph Neural Network (GNN) is a powerful deep learning method and have achieved great success in prediction tasks on graph-structured data \cite{wu2020comprehensive}. In general, GNN models can be categorized into two types: spatial GNNs including message-passing neural networks (MPNN) \cite{gilmer2017neural}, graph attention networks (GAT) \cite{velivckovic2017graph}, graph isomorphic networks (GIN) \cite{xu2018powerful}, which propagate node neighbouring information and update  node representations via weighted sum or the average of the neighbours, spectral GNNs such as ChebyNet \cite{defferrard2016convolutional}, GCN \cite{kipf2016semi}, BernNet \cite{he2021bernnet}, which involve filtering mechanism in the graph spectral domain generated from graph Fourier transform where the orthonormal system is constructed by the eigenvectors of the graph Laplacian. As one class of spectral models, the graph wavelet representing \cite{yao2022sparse} provides a multi-resolution analysis of graph signals to capture the node information via different scales and usually produces a better signal representations due to the reconstructable (known as \textit{tightness}) property of signal decomposition. In particular, graph wavelet frames (known as graph framelets) effectively separates the low and high frequency signal information and has been developed as graph convolution models in the work \cite{zheng2022decimated}.

In recent years, apart from developing more advanced GNNs, the GNN's learning theory on both of its expressive power and limitations has also attracted great attention. Several pitfalls of GNN have been identified in terms of its computational aspects such as over-smoothing \cite{cai2020note} and over-squashing \cite{topping2021understanding}. Theoretical aspects such as limited expressive power \cite{xu2018powerful} and adaptation to heterophily graph dataset \cite{zheng2022semi} have also been studied intensively. To address these issues, many attempts have been made by applying the techniques in dynamic systems and differential equations \cite{di2022graph} and algebraic topology \cite{bodnar2022neural}. In particular, graph geometric features, known for their ability to capture the intrinsic geometric properties of graphs, have been effectively employed to enhance the representation of graphs \cite{feng2023community}. By incorporating these graph geometric features, GNNs can propagate node feature information based on the underlying geometric characteristics. This utilization leads to improved learning outcomes and enhanced robustness against potential learning challenges such as over-smoothing.


Despite the remarkable prediction accuracy achieved by framelet models in various graph learning tasks, it remains unclear whether these models fully leverage the intrinsic geometric and topological information of graphs to address the aforementioned challenges. To address this gap, our paper aims to enhance the graph framelet model by incorporating graph Ricci curvature, allowing for a comprehensive integration of graph geometric information. Inspired by curvature in the continuous domain, graph Ricci curvature \cite{ollivier2007ricci} quantifies the deviation of the geometry between pairs of neighborhoods from a 'flat' case, such as a grid graph. Integrating graph curvature information enables a more geometrically informed design of graph convolutions, leading to substantial improvements in model performance. Notably, we demonstrate that by incorporating the graph adjacency and curvature ($\kappa$) information using a carefully selected transformation function $\zeta(\kappa)$, the enhanced framelet model becomes capable of effectively handling both homophily and heterophily graph datasets. To the best of our knowledge, this work provides the first theoretical support for the empowering role of graph Ricci curvature in multi-resolution graph neural networks. We summarize our contributions as follows:

\begin{itemize}
    \item We establish a connection between graph intrinsic topological information (graph Ricci curvature) and multi-resolution graph neural networks, enabling the propagation of edge connectivity importance through both low and high-frequency paths.
    
    \item We introduce graph geometry-based criteria for selecting the appropriate functional ($\zeta$) that operates on the graph Ricci curvature. This ensures that the curvature information is effectively captured by the graph neural network while preserving crucial graph properties, such as the graph Laplacian.
    
    \item We present two curvature-based framelet models, namely RC-UFG (Hom) and RC-UFG (Het), showcasing their enhanced adaptability on homophilic and heterophilic graph datasets from an energy dynamic perspective. Furthermore, drawing inspiration from the relationship between Ricci curvature and graph topology, we develop a curvature-based graph edge drop (CBED) technique and establish its equivalence with RC-UFG (Het). 
    
    \item We thoroughly evaluate the performance of the proposed models on graph learning tasks involving both homophilic and heterophilic graphs. The experimental results demonstrate the effectiveness of our models in real-world node classification tasks.
\end{itemize}

The remainder of the paper is organized as follows. Section \ref{related_works} reviews the literature on graph convolution, graph Ricci curvature, and dynamic systems. We also include an overview of graph framelet transform and convolution. Section \ref{Preliminaries} includes some preliminary formulations on graph, graph convolution and graph Ricci curvature. In section \ref{ricci_lap}, we show how curvature framelet convolution is built by providing the criteria of selecting the functional $\zeta$ onto graph Ricci curvature based on the analogy between the notions of Ricci curvature in manifold (smooth) and graph (discrete) settings. We present the theoretical support on the benefits of selecting such form of $\zeta$ from energy dynamic perspective. Furthermore, we develop a curvature based graph edge drop algorithm (CBED) and show the link between CBED and $\zeta$. Section \ref{experiment} provides comprehensive experiments on comparing our proposed model with baseline models on homophilic and heterophilic graph datasets to show its superior prediction power. Finally, this paper is concluded in Section \ref{conclusion}.

\section{Related Works}
\label{related_works}
\subsection{Graph convolution, dynamic system}
In the past decades, researchers have been working on how to conduct convolutive operations on graphs. GNNs are a type of deep learning architecture based model that can leverage the graph structure and aggregate node information from the neighborhoods in a convolutional fashion and have achieved notable successes in various graph representation and prediction tasks. One major research direction is to define graph convolution from spectral aspect, thus the so-called graph wavelets \cite{zheng2021framelets} gradually gain its popularity. Another direction for graph convolution research is conducted based on spatial (node) information of the graph. One common process within most of spatial methods is to aggregate node representations from its  neighbourhood  \cite{gilmer2017neural,kipf2016semi}. 

One active routine for exploring the behavior of GNNs is through continuous dynamic systems and differential equations. For example, the graph convolution network (GCN) \cite{kipf2016semi} can be seen as an analogy to a discrete Markov random walk with its transition matrix determined by the random walk graph Laplacian \cite{oono2019graph}. Furthermore, the linearized version of GCN is verified as the discrete heat diffusion which minimizes the graph Dirichlet energy. More recently,\cite{di2022graph} introduces a framework for analyzing the energy evolution of GNNs in terms of gradient flow. Specifically, the work proposes a general energy where its gradient flow leads to many existing GNN models. Under such framework, \cite{di2022graph} verifies that many models can only lead to low-frequency-dominant (LFD) dynamics, including GCN \cite{kipf2016semi}, GRAND \cite{chamberlain2021grand} and CGNN \cite{xhonneux2020continuous}. In addition, the discussion on the evolution of energy dynamic on the graph framelet is established recently by \cite{han2022generalized}.
\subsection{Graph Ricci curvature}
The notion of graph Ricci curvature on general spaces without Riemannian structures has been recently studied \cite{ollivier2007ricci}. One path of defining graph curvature is through Forman's discretization on the Ricci curvature defined on polyhedral or the CW complexes \cite{forman2003bochner}. Although this type of curvature was proposed relatively recently, there have already been
a number of papers investigating properties of these measures and applying them to real-world graphs/networks  \cite{das2018effect}. Another way to define graph curvature is through Ollivier's discretization which was first explored in \cite{ollivier2007ricci}. Both Ollivier–Ricci curvature and Forman-Ricci curvature assign measure, i.e. a real number, 
to each edge of the given network, but they are calculated in quite different ways by 
capturing different metric properties of a Riemannian manifold. The Ricci curvature based on optimal transportation theory has become a popular topic in various fields such as identifying tumor-related genes from normal genes \cite{sandhu2015graph}, predicting and managing the financial market risks \cite{sandhu2016Ricci}. In terms of GNNs, graph Ricci curvature has been considered to enhance the capacity of GNNs. For example \cite{di2022over} presented a curvature based framework to describe both over-squashing and over-smoothing issues via a multi-particle dynamic framework.

\subsection{Graph framelet and transforms}
Wavelet analysis on graph structured data was first explored in \cite{crovella2003graph} in which polynomials of a differential operator were used to build multi-scale transforms. Recently \cite{behjat2016signal} established energy spectral density to form tight frames on a graph by considering both graph topological and signal features. In machine learning,  Harr-like orthonormal wavelet system, first explored by \cite {chui2015representation}, has been applied to deep learning models for undirected graphs \cite{zheng2020mathnet}. Recent work has shown that framelet transform based graph convolution is capable of greatly improving the GNN learning outcomes \cite{zheng2021framelets,lin2023magnetic}, and framelet regularizer has been applied to graph denoising tasks \cite{zhou2021graph}. In addition, \cite{chendirichlet} proposed a spatial graph framelet convolution and showed its close connection to Dirichlet energy.



\section{Preliminaries}
\label{Preliminaries}
\subsection{Graph and graph convolution}
We denote a graph $\mathcal G = (\mathcal V_\mathcal G,\mathcal E_\mathcal G)$ where $\mathcal V_\mathcal G$ and $\mathcal E_\mathcal G$ represent the sets of vertices and edges, respectively. We also consider $\mathbf X=[x_1^\top;\ldots;x_n^\top] \in \mathbb R^{n \times d_0}$ as the feature matrix of the $n$ nodes with each node feature vector $x_i\in \mathbb R^{d_0}$. For any distinct node pairs $x_i ,x_j \in \mathcal V_\mathcal G$ we denote $i \sim j$ if they are connected with an edge. For any finite graph $\mathcal G$, its normalized Laplacian $\widehat{\mathbf L } = \widetilde{\Delta} = \mathbf I_n - \mathbf {\widehat{A}} \succeq 0$ is a positive semi-definite matrix, where $\widehat{\mathbf A} =\mathbf D^{-\frac12}  \widetilde{\mathbf A} \mathbf D^{-\frac12}$is the normalized adjacency matrix and $\widetilde{\mathbf A} = \mathbf A+\mathbf I_n$. In addition, for any node $i$ of $\mathcal G$, we will let $d_i$ be its degree. Furthermore, given a symmetric matrix $\mathbf B$, we let $\rho_\mathbf B$ be its spectral radius.


Graph convolution network (GCN) \cite{kipf2016semi} defines the layer-wise propagation rule via the normalized adjacency matrix as
\begin{equation}
    \mathbf H(\ell + 1) = \sigma \big( \widehat{\mathbf A} \mathbf H(\ell) \mathbf W^\ell  \big), \label{eq_classic_gcn}
\end{equation}
where $\mathbf H(\ell)$ denotes the feature matrix at layer $\ell$ with $\mathbf H(0) = \mathbf X \in \mathbb R^{n \times d_0}$, i.e. the input signals, and $\mathbf W^\ell$ is the learnable feature transformation. It is easy to verified that GCN corresponds to a localized filter by the graph Fourier transform, i.e., $\mathbf h(\ell+1) = \mathbf U^\top (\mathbf I_n - \mathbf \Lambda) \mathbf U \mathbf h(\ell)$,where $\mathbf U, \mathbf \Lambda$ are from the eigendecomposition $\widehat{\mathbf L} = \mathbf U^\top \mathbf \Lambda \mathbf U$ and $\mathbf U \mathbf h$ is known as the Fourier transform of a graph signal $\mathbf h \in \mathbb R^n$. Let $\{ (\lambda_i, \mathbf u_i) \}_{i=1}^n$ be the set of eigenvalue and eigenvector pairs of $\widehat{\mathbf L}$. Based on the spectral graph theory \cite{chung1997spectral}, we have $\rho_{\widehat{\mathbf L}} \leq 2 $, and the equality only holds if there exists a connect component of the graph that is bipartite.

\subsection{Graph framelet convolution}

Graph (undecimated) framelets \cite{zheng2022decimated} are defined by a filter bank $\eta_{a,b} = \{ a; b^{(1)}, ..., b^{(L)} \}$ and its induced (complex-valued) scaling functions $\Psi = \{ {\alpha}; \beta^{(1)}, ..., \beta^{(L)} \}$ where $L$ represents the number of high-pass filters.
Particularly, the relationship between the filter bank and scaling functions is:
\begin{align*}
\widehat{\alpha}(2\xi) = \widehat{a}(\xi) \widehat{\alpha}(\xi)   \,\,\, \text{and} \,\,\, &\widehat{\beta^{(r)}}(2\xi) = \widehat{b^{(r)}}(\xi) \widehat{\alpha}(\xi) \\
& \,\,\, \forall \xi \in \mathbb R, r =1,...,L
\end{align*}
where  $\widehat{\alpha}, \widehat{\beta^{(r)}}$ are the Fourier transformation of $\alpha,\beta^{(r)}$, and $\widehat{a}, \widehat{b^{(r)}}$ are  the corresponding Fourier series of $a, b^{(r)}$ respectively.  The graph framelets are then defined  by $\varphi_{j,p}(v) = \sum_{i=1}^n \widehat{\alpha}\big( {\lambda_i}/{2^j} \big) u_i(p) u_i(v)$ and $\psi^r_{j,p}(v) = \sum_{i = 1}^n \widehat{\beta^{(r)}} \big( \lambda_i/ 2^j \big) u_i(p) u_i(v)$ for $r = 1,..., L$ and for scale level $j = 1,...,J$. We use $u_i(v)$ to represent the eigenvector $\mathbf u_i$ at node $v$. $\varphi_{j,p}$ and $\psi^r_{j,p}$ are known as the \textit{low-pass framelets} and \textit{high-pass framelets} at node $p$. 

The \textit{framelet coefficients} of a graph signal $\mathbf h$ are given by $\mathbf v_{0} = \{ \langle \mathbf {\varphi}_{0,p} , \mathbf h \rangle \}_{p \in \mathcal V_\mathcal G}$, $\mathbf w_{j}^r = \{ \langle \mathbf \psi_{j,p}^r, \mathbf h \rangle \}_{p \in \mathcal V_\mathcal G}$. For a multi-channel signal $\mathbf H \in \mathbb R^{n \times d_0}$, we have its framelet coefficients as
\begin{align*}
    &\mathbf V_0 = \mathbf U^\top \widehat{\alpha} \Big( \frac{\mathbf \Lambda}{2} \Big) \mathbf U \mathbf H, \\
    &\mathbf W_j^r = \mathbf U^\top \widehat{\beta^{(r)}} \Big( \frac{\mathbf \Lambda}{2^{j+1}} \Big) \mathbf U \mathbf H,\ 
    \forall j=1,...,J, r = 1,...,L.
\end{align*}
Write $\mathbf \Lambda_{0,J}=\widehat{\alpha} \Big( \frac{\mathbf \Lambda}{2} \Big)$ and $\mathbf \Lambda_{r,j}=\widehat{\beta^{(r)}} \Big( \frac{\mathbf \Lambda}{2^{j+1}} \Big)$. 
From tightness of the framelet transform, we have $\mathbf \Lambda_{0,J} + \sum_{r,j} \mathbf \Lambda_{r,j} = \mathbf I_n$. 
Furthermore, let us define  the framelet transformation matrices $\mathcal W_{0,J}$, $\mathcal W_{r,J}$ such that: 
\begin{align*}
     &\mathcal W_{0,J} = \mathbf U^\top (\mathbf \Lambda_{0,J})\mathbf U  \\  
     &\mathcal W_{r,J}  =  \mathbf U^\top  (\mathbf \Lambda_{r,J}) \mathbf U ,\ \forall j = 1,...,J, r = 1,...,L.
\end{align*}
Since framelet decomposition and reconstruction are invertible to each other, thus we have $\mathcal W_{0, J}^\top \mathcal W_{0,J} \mathbf H + \sum_{r,j} \mathcal W_{r,j}^\top \mathcal W_{r,j} \mathbf H = \mathbf H$. There are in general two types of graph framelets. The first type is \textit{spectral framelet} proposed in \cite{yang2022quasi} where the filter functions are applied in the graph frequency domain before reconstructing the signals. Thus the node feature propagation rule for \textit{spectral framelet} is: 
\begin{align}
\mathbf H(\ell + 1) &= \sigma( \mathcal W_{0,J}^\top {\rm diag}(\mathbf \theta_{0,J}) \mathcal W_{0,J} \mathbf H(\ell) \mathbf W^\ell +
\sum_{r,j} \mathcal W_{r,j}^\top {\rm diag}(\mathbf \theta_{r,j}) \mathcal W_{r,j} \mathbf H(\ell) \mathbf W^\ell),
\end{align}
where $\boldsymbol{\theta} \in \mathbb R^n$ contains learnable coefficients in each high or low frequency domain and $\mathbf W^\ell$ is a shared weight matrix. Instead of performing spectral filtering, the \textit{spatial framelet} developed in \cite{chendirichlet}, the second type, performs a spatial message passing over the spectral framelet domain and thus yields an adjacency information based propagation rule as:
\begin{align}
\mathbf H(\ell + 1) & = \mathcal W^T_{0,J} \sigma( \widehat{\mathbf A} \mathcal W_{0,J} \mathbf H(\ell) \mathbf W^\ell_{0,J}) + \sum_{r,j} \mathcal W_{r,j}^\top \sigma( \widehat{\mathbf A} \mathcal W_{r,j} \mathbf H(\ell) \mathbf W^\ell_{r,j})). 
\end{align}
In this paper, our subsequent analysis are mainly focus on the {spectral framelet} or just framelet for simplicity, although we will also provide some outcomes for {spatial framelet} as byproduct in later sections. Furthermore, to alleviate the computational cost from eigendecomposition of the graph Laplacian, followed by the work in \cite{dong2017sparse}, we approximate the framelet transformation matrix by the Chebyshev polynomials with certain degree $n$. For notation simplicity, in the sequel, we use  $\mathcal{T}_j(\xi)$ instead of $\mathcal{T}^n_j(\xi)$. Then the Framelet transformation matrices can be approximated as:
\begin{align}
    \mathcal{W}_{0,J} &\approx
    \mathcal{T}_0(\frac1{2^{L+m}}\widetilde{\mathbf L}) \cdots \mathcal{T}_0(\frac{1}{2^{m}}\widetilde{\mathbf L}), 
    \;\;\;\;\;\; \\
    \mathcal{W}_{r,0} &\approx 
    \mathcal{T}_r(\frac1{2^{m}}\widetilde{\mathbf L}),  \;\;\; \text{for } r = 1, ..., L, \label{eq:Ta}
 \\ 
    \mathcal{W}_{r,j} &\approx 
    \mathcal{T}_r(\frac{1}{2^{m+\ell}}\widetilde{\mathbf L})\mathcal{T}_0(\frac{1}{2^{m+\ell-1}}\widetilde{\mathbf L}) \cdots \mathcal{T}_0(\frac{1}{2^{m}}\widetilde{\mathbf L})\;\;\;\; \\
    \;\;\;\;\text{for } r &=1, ..., L, j = 1, ..., J.\notag
\end{align}

\subsection{Graph Ricci curvature and Ricci flow} \label{ricci_define}
\begin{figure}[t]
     \centering
     \scalebox{1.2}{
     \subfloat[]{\includegraphics[width =0.22\textwidth, height=0.20\textwidth]{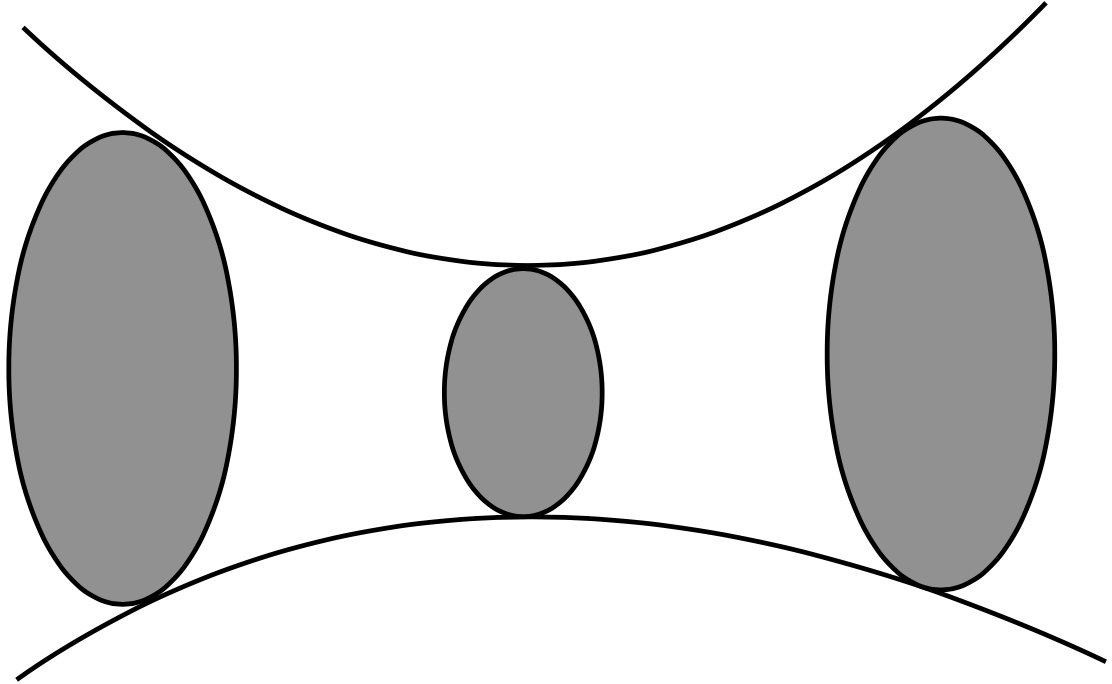}}\;\;\;
     \subfloat[]{\includegraphics[width =0.22\textwidth, height = 0.20\textwidth]{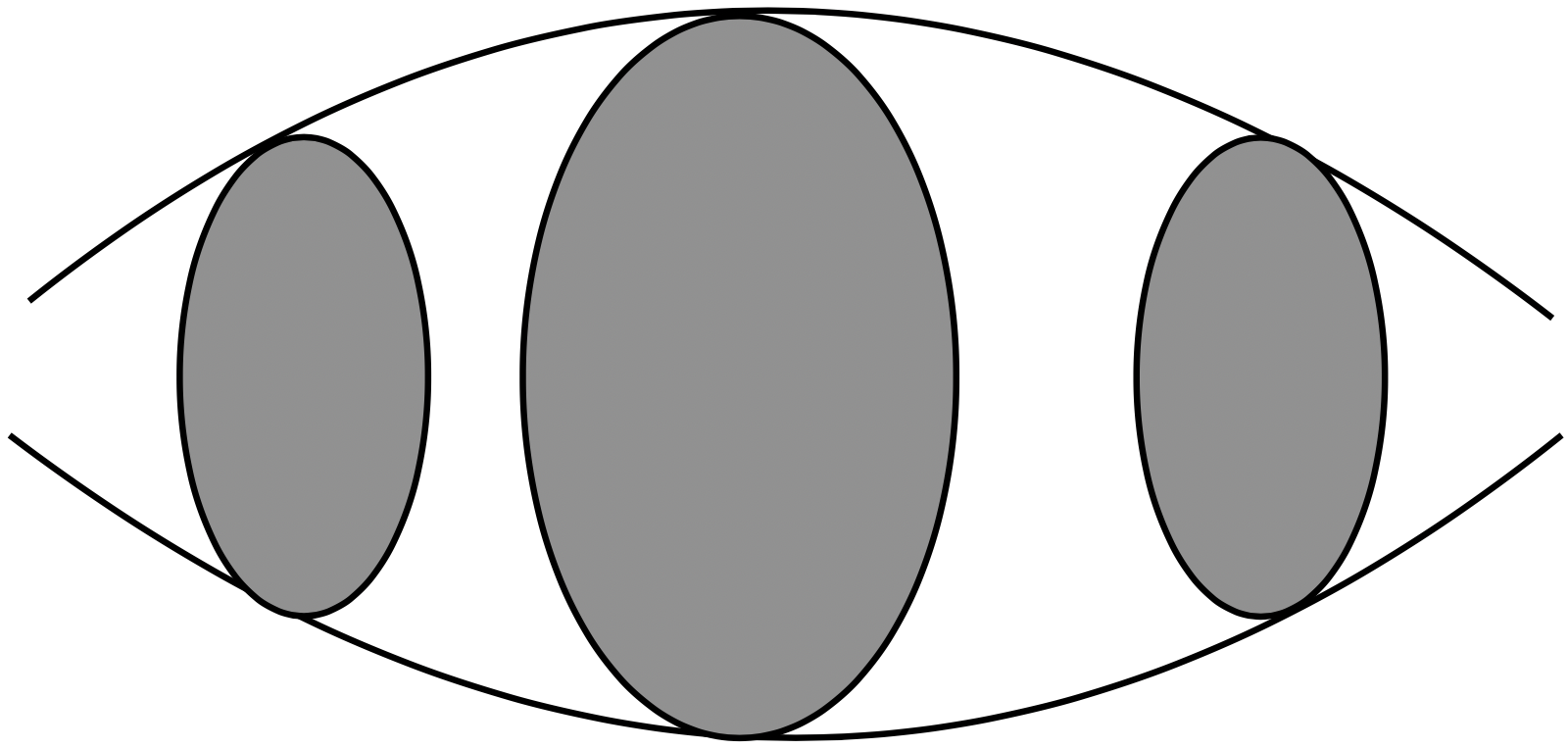}}\;\;\;}
\caption{The geometric interpretation of Ricci curvature. Ricci curvature measures the changes of the volume along with different geodesic directions. It also shows the dispersion rate of the family of geodesics with the same initial point. 
Figure 1(a) shows in the hyperbolic space two parallel geodesics tend to disperse with each other, where the sectional curvature is negative, and Figure 1(b) shows in the spherical space two geodesics tend to converge together in opposite case.} 
\label{ricci_smooth}
\end{figure}

In this section, we provide the formulation of graph (Ollivier) Ricci curvature and Ricci flow. We first provide an intuitive description of the Ricci curvature defined on a smooth Riemannian manifold as follows. Let $\mathcal M$ be a Riemannian manifold. $x,y \in \mathcal M$ are two close points on $\mathcal M$. We generate two parallel geodesics passing through $x$ and $y$ as the following. Pick a tangent vector $\mathbf w$ on $x$ from which the geodesic will not reach $y$. Parallel transport $\mathbf w$ to $y$ as $\mathbf w'$. Then generate two parallel geodesics along $\mathbf w $ and $\mathbf w'$ from $x$ and $y$ respectively, called $g_1$ and $g_2$. So $g_1$ and $g_2$ has no acceleration to each other on $\mathcal M$. 
$g_1$ and $g_2$ will converge to each other if the curvature (known as sectional curvature) is positive, and will diverge away if it is negative. When the manifold is locally flat, two geodesics will never meet. Furthermore, if we consider all the possible directions of $\mathbf w$ from $x$ to $\mathbf w'$, then the average of all sectional curvature defines the Ricci curvature. Figure \ref{ricci_smooth} shows a general view of Ricci curvature visualized as an ellipsoid traveling between the geodesics and volume (as shaded area) changes due to different curvatures. 

In terms of the graph setting, we focus on the Ollivier Ricci curvature  \cite{ollivier2007ricci} on graph, which is an {\it edge-based} curvature. Before we introduce the definition, we define a probability measure at node $i \in \mathcal V_\mathcal G$ for a given $\alpha\in [0,1]$
\begin{equation*}
    m_i(j) = 
    \begin{cases}
        \frac{1}{|\mathcal{N}_i|}, & j \in \mathcal{N}_i \\
        0, & \text{ otherwise}
    \end{cases}
\end{equation*}
where $|\mathcal{N}_i|$ is the size of ${\mathcal N}_i$, i.e. the degree of $x_i$. We highlight that this is the original definition in \cite{ollivier2007ricci} and there exist many alternatives to define $m_i$ as long as each $m_i$ generates a discrete distribution over every node in $V$. Then the graph Ollivier Ricci curvature between node $i,j$ is defined as
\begin{equation}\label{ricci_defin}
    \kappa(i,j) = 1 - \frac{W_1(i,j)}{d_s(i,j)},
\end{equation}
where $d_s(i,j)$ is the shortest path distance on $\mathcal{G}$ between nodes $i, j$ and $W_1(i,j)$ is the $L_1$-Wasserstein distance computed as $W_1(i,j) = \inf_{\Gamma} \sum_{i'} \sum_{j'} \Gamma_{i'j'} d_s(i',j')$ where $\Gamma$ is the joint distribution satisfies the coupling conditions, i.e., $\sum_{i'}\Gamma_{i'j'} = m_j(j')$, $\sum_{j'} \Gamma_{i'j'} = m_i(i')$ for all $i',j'$. Also, it is obvious to see that the range of Ricci curvature is between $-2$ and $1$.

In addition to above formulation of Ricci curvature, the graph Ricci flow method \cite{ni2019community} is an analogy of the Ricci flow approach from the continuous domain.
Specifically, let $a_{i,j}$ represents the weight between nodes $i$ and $j$, for community detection, the Ricci flow method update the edge weights iteratively as: 
\begin{align}\label{ricci_flow}
    a_{i,j} = d_s{(i,j)} (1 - \kappa(i,j))
\end{align}
where both $d_s{(i,j)}$ and $\kappa(i,j)$ 
are calculated using the weight $a_{i,j}$ at current iteration (i.e., $d_s{(i,j)}$ can be obtained from Dijkstra's algorithm).Thus the process can be interpreted as increasing the edge weight for negatively curved edges and decreasing the edge weight for positively curved ones. In the smooth (manifold) setting, the Ricci flow method provides a continuous shape (metric) deformation to the manifold at the same time preserving its geometric structures. Similar to the discrete (graph) setting, one can interpret such method as a step-wise evolution of connectivity importance (strength) of the graph while preserving its adjacency information.

\subsection{Graph Dirichlet energy, gradient flow and consistency}
The graph Dirichlet energy is a measure of presenting the total variations for node features, which is defined as follows. 
\begin{defn}[Graph Dirichlet Energy]\label{Dirichlet_energy}
Given signal embedding matrix $\mathbf H{(\ell)} = :\{\mathbf h_1{(\ell)},\mathbf h_2{(\ell)}...\mathbf h_N{(\ell)}\}^T \in \mathbb R^{n\times d_\ell}$ learned from GCN at the $\ell$-th layer, the Dirichlet energy $E(\mathbf H{(\ell)})$ is defined as: 
\begin{align*}
    E(\mathbf H{(\ell)}) &= \mathrm{Tr}(\mathbf H{(\ell)^T}\widetilde{\Delta}\mathbf H{(\ell)}) =\frac{1}{2}\sum_{i,j}a_{i,j}\|\frac{\mathbf h_i{(\ell)}}{\sqrt{1+d_i}}-\frac{\mathbf h_j{(\ell)}}{\sqrt{1+d_j}}\|^2_2.
\end{align*}
\end{defn}
The gradient flow of the Dirichlet energy yields the so-called graph heat equation \cite{chung1997spectral} as $\dot{\mathbf H}(t) = - \nabla E(\mathbf H(t)) = - \widehat{\mathbf L} \mathbf H(t)$. Its Euler discretization leads to the propagation of linear GNN models. The process is called Laplacian smoothing and it converges to the kernel of $\widehat{\mathbf L}$, i.e., ${\rm ker}(\widehat{\mathbf L})$, as $t \xrightarrow{} \infty$, resulting in non-separation of nodes with same degrees, known as the over-smoothing issue. 

Recently, the work done by \cite{di2022graph} shows that even with non-linear activation and weight matrix in GCN \eqref{eq_classic_gcn}, the dynamics of the model are still  dominated by the low frequency information and eventually converge to the kernel of  $\widehat{\mathbf L}$. In this paper, we refer to \cite{di2022graph} to quantify this phenomenon in terms of the gradient flow of Dirichlet energy. Specifically, consider a general dynamic as $\dot{\mathbf H}(t) = {\rm GNN}_\theta (\mathbf H(t), t)$, with ${\rm GNN}_\theta(\cdot)$ as an arbitrary graph neural network function. We can define GNNs low frequency and high frequency dominant behavior as follows: 

\begin{defn}[\cite{di2022graph}]
\label{def_hfd_lfd}
$\dot{\mathbf H}(t) = {\rm GNN}_\theta (\mathbf H(t), t)$ is Low-Frequency-Dominant (LFD) if $E \big(\mathbf H(t)/ \| \mathbf H(t) \| \big) \xrightarrow{} 0$ as $t \xrightarrow{} \infty$, and is High-Frequency-Dominant (HFD) if $E \big(\mathbf H(t)/ \| \mathbf H(t) \| \big) \xrightarrow{} \rho_{\widehat{L}}/2$ as $t \xrightarrow{} \infty$. 
\end{defn}

\begin{lem}[\cite{di2022graph}]
\label{lemma_hfd_lfd}
A GNN model is LFD (resp. HFD) if and only if for each $t_j \xrightarrow{} \infty$, there exists a sub-sequence indexed by $t_{j_k} \xrightarrow{} \infty$ and $\mathbf H_{\infty}$ such that $\mathbf H(t_{j_k})/\| \mathbf H(t_{j_k})\| \xrightarrow{} \mathbf H_{\infty}$ and $\widehat{\mathbf L} \mathbf H_{\infty} = 0$ (resp. $\widehat{\mathbf L} \mathbf H_{\infty} = \rho_{\widehat{L}} \mathbf H_{\infty}$).
\end{lem}

The characterization of LFD and HFD has direct consequences on model performance for graphs with different level of consistencies, known as homophily and heterophily, which can be defined as: 
\begin{defn}[Homophily and Heterophily \cite{FuZhaoBian2022}]
\label{HomophilyHeterophily}
The homophily or heterophily of a network is used to define the relationship between labels of connected nodes. The level
of homophily of a graph can be measured by $\mathcal{H(G)} = \mathbb{E}_{i \in \mathcal{V}}[|\{j\}_{j \in \mathcal{N}_{i,y_j= y_i}}|/|\mathcal{N}_i|]$, where 
$|\{j\}_{j \in \mathcal{N}_{i,y_j= y_i}}|$ denotes the number
of neighbors of $i \in V$ that share the same label as $i$ such that $y_i = y_j$, and $\mathcal{H(G)} \rightarrow 1$ corresponds to strong homophily
while $\mathcal{H(G)} \rightarrow 0$ indicates strong heterophily. We say that
a graph is a homophilic (heterophilic) graph if it has strong homophily (heterophily). 
\end{defn}

\begin{rem}[LFD, HFD and graph homophily]
\label{challenge}
If $\mathcal G$ is homophilic, adjacent nodes are likely to share the same label and we expect a smoothing learning dynamic induced from the graph learning model. In the opposite case of heterophily, in which identical labels of one specific node are likely contained in the distant nodes, and in this case we expect to have a graph learning model that can capture such pattern by inducing a sharpening learning dynamic on the graph rather than smoothing everything out. As shown in Lemma \ref{lemma_hfd_lfd}, a GNN model is LFD if and only if there exists a sub-sequence of $\mathbf H(t)$ that converges to ${\rm ker}(\widehat{\mathbf L})$ (thus smoothing) and is HFD if and only if there exists a sub-sequence that converges to the eigenvector associated with the largest frequency of $\widehat{\mathbf L}$ (thus separating). In terms of graph framelet, if the graph is homophily, one shall let the smoothing effect in the low frequency domain dominates the one in the high frequency frame domain, and on the other hand, once the graph is highly heterophily, one shall expect the evolution in the high frequency domain dominates the one in the low frequency domain. In the next section we show this task can be done by inserting curvature information to the adjacency matrices in both low and high frequency domains. 
\end{rem}

\section{Curvature Framelet Convolution} \label{ricci_lap}
\begin{figure*}[t]
    \centering
    \includegraphics[width = 0.9
    \textwidth, height = 0.55\textwidth]{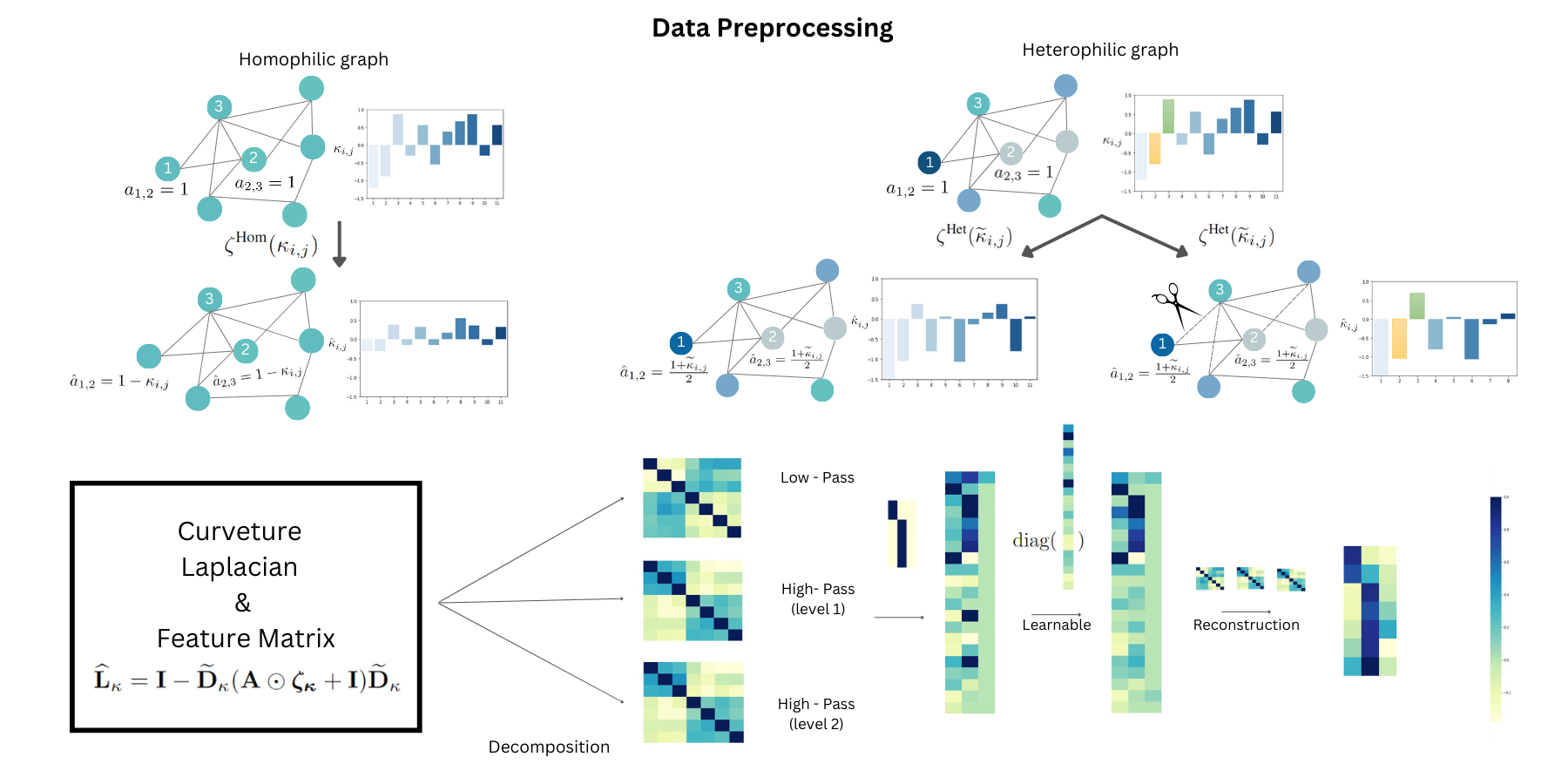}
    \caption{Illustration of the working process in curvature enhanced framelet convolution (RC-UFG (Hom) and RC-UFG (Het)). The upper half of the figure shows three type of curvature based data pre-processing methods including $\zeta^\text{Hom}$, $\zeta^\text{Het}$ and curvature based edge dropping (CBED) to enhance framelet's adaptation on homophilic and heterophilic graphs. 
    Specifically, the functionality of $\zeta^\text{Hom}$ is to shrink the curvature distribution of the graph (upper left) so that a stronger LFD dynamic can be generated in framelet (Lemma \ref{lap_eigen_range}); whereas both $\zeta^\text{Het}$ and CBED are capable of reducing the minimum curvature and resulting a stronger HFD dynamic of framelet (Lemma \ref{lower_bound} and description of CBED in Section \ref{sec:CBED}). After the data pre-processing, the curvature information based graph Laplacian $\widehat{\mathbf{L}}_\kappa$ is then utilized together with feature matrix for framelet system to generate the learning outcomes.}
    \label{framelet_structure}. 
\end{figure*}

In this section, we show how the curvature information can be inserted into the graph framelets by a carefully selected transformation $\zeta$. Recall that without considering the activation function, the information propagation of \textit{spectral framelet} can be written as: 
\begin{align}
\label{spectral_framelet}
\mathbf H(\ell + 1) &= \mathcal W_{0,J}^\top {\rm diag}(\mathbf \theta_{0,J}) \mathcal W_{0,J} \mathbf H(\ell) \mathbf W^\ell + \sum_{r,j} \mathcal W_{r,j}^\top {\rm diag}(\mathbf \theta_{r,j}) \mathcal W_{r,j} \mathbf H(\ell) \mathbf W^\ell,
\end{align}
Due to the homophily and heterophily property differences  between graphs, the way of inserting curvature information to framelet should be different. However, it is admitted that for any finite undirect graph $\mathcal G = (\mathcal V_\mathcal G,\mathcal E_\mathcal G)$, its normalized Laplacian $\widehat{\mathbf L} = \widetilde{\Delta} = \mathbf I_n - \mathbf {\widehat{A}}\succeq 0$ is a positive semi-definite (SPD) matrix, and based on the spectral graph theory \cite{chung1997spectral}, $\rho_{\widehat{\mathbf L}} \leq 2$ and the equality holds if and only if there exists a connected component of the graph (or subgraph) $\mathcal G$ that is bipartite. Therefore, to have $\widehat{ \mathbf L }\succeq 0$ we must have $a_{i,j} \geq 0 \, \forall i \sim j$. However, since the range of Ricci curvature can be negative, a transformation $\zeta(\kappa_{i,j}) \rightarrow \mathbb R_+$ is needed. Therefore, we summarize the basic conditions of $\zeta$ based on the above reasoning as follows:  
\begin{enumerate}
    \item the target domain of the transformation $\zeta$ must be non-negative and hence $\widehat{\mathbf L }$ remains SPD;
    \item the transformation $\zeta$ must be injective (one-to-one) to ensure no information loss; 
\end{enumerate}

Additionally to the basic conditions on $\zeta$, in the rest of this section, we will show in Section \ref{basisc_condition} the form of $\zeta^\text{Hom}$  which inserts the curvature information to framelet to enhance its homophily adaption. In Section \ref{heterophily_adaption} we show the form of $\zeta^\text{Het}$ which enhances the model to fit the heterophily graph input. In addition, inspired by the relationship between graph Ricci curvature and graph topology, we will also develop a curvature based graph edge drop (CBED) and prove that the framelet with $\zeta^\text{Het}$ is a kind of soft version of CBED. We summarize the working process of our curvature based framelet model in Fig.~\ref{framelet_structure}.

\subsection{Conditions on $\zeta$ when the input graph is homophilic}
\label{basisc_condition}
There are various choices of $\zeta$ based on the basic conditions. However, as we have mentioned in Remark \ref{challenge}, when the input graph is homophilic, an ideal GNN shall be able to induce a smoothing dynamic that is higher than its sharpening counterpart. To achieve this goal, recall that the graph Ricci flow \cite{ni2019community} updates the edge weight (treated as the discrete metric) with the following equation: 
\begin{align*}
    a_{i,j} = d_s{(i,j)} (1 - \kappa_{i,j}),
\end{align*}
where we use $\kappa_{i,j}$ instead of $\kappa({i,j})$ for a simpler notation.
When the graph is unweighted, we have  $d_s{(i,j)} = a_{i,j} = 1$, therefore the first iteration of the Ricci curvature is $w_{i,j}^{+} = 1 - \kappa_{i,j}$. Therefore, the curvature based edge weights induce from Ricci flow is: 
\begin{align}\label{zeta_homo}
    \zeta^\text{Hom}(\kappa_{i,j}) = d_s{(i,j)} (1 - \kappa_{i,j}) = 1 -\kappa_{i,j}.
\end{align}
\begin{figure}[H]
     \centering
     \subfloat[]{\includegraphics[width = 0.3\textwidth, height = 0.18\textwidth]{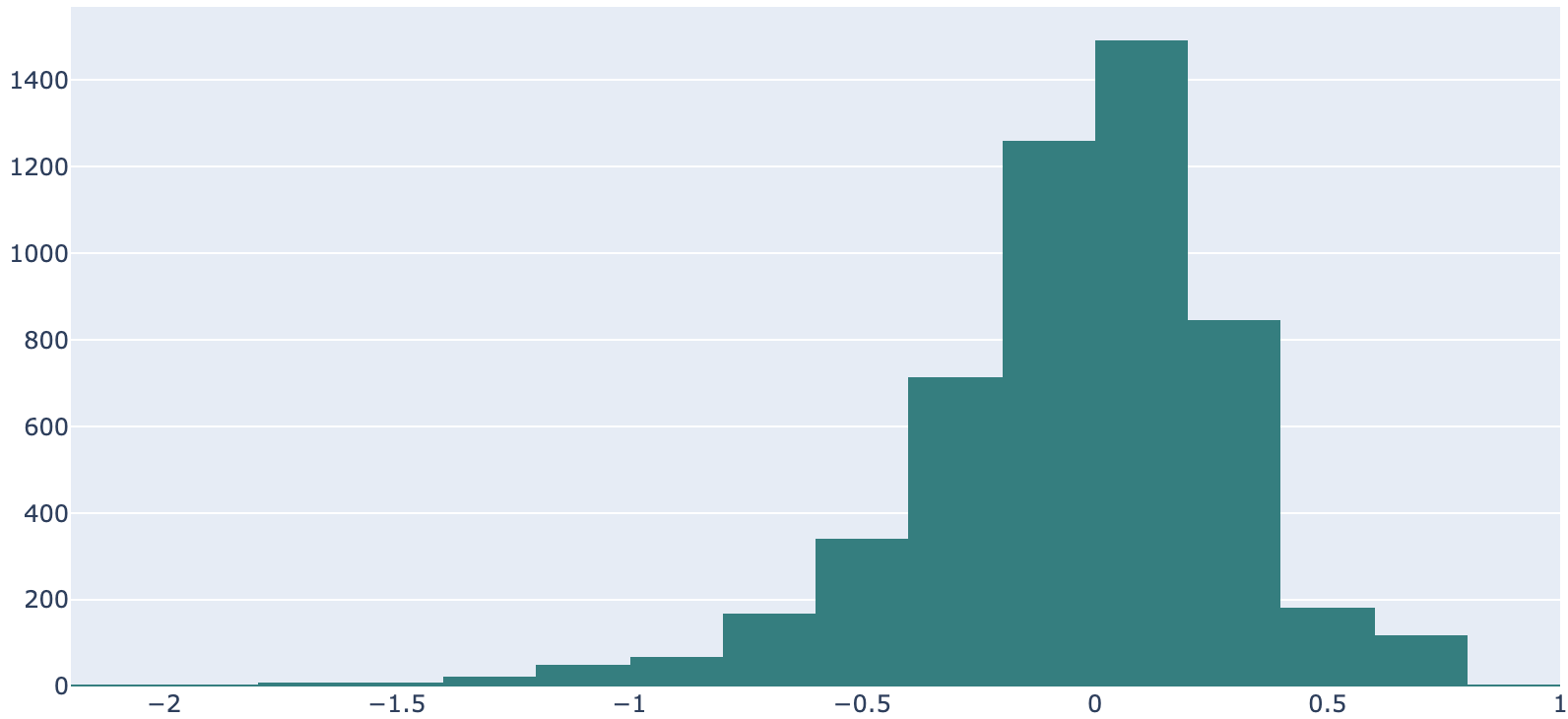}} \;\;\;
     \subfloat[]{\includegraphics[width =0.3\textwidth, height = 0.18\textwidth]{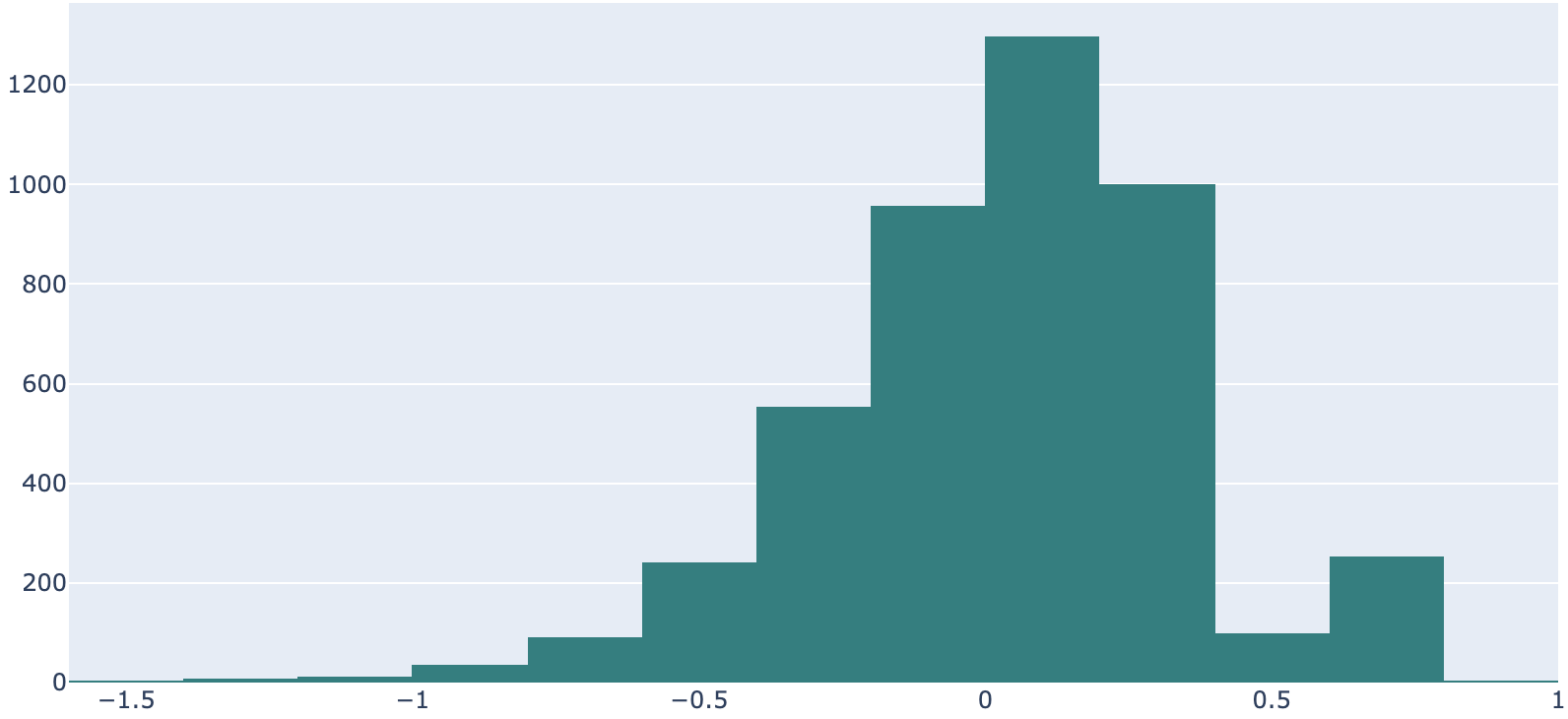}}\;\;\;
     \subfloat[]{\includegraphics[width = 0.3\textwidth, height = 0.18\textwidth]{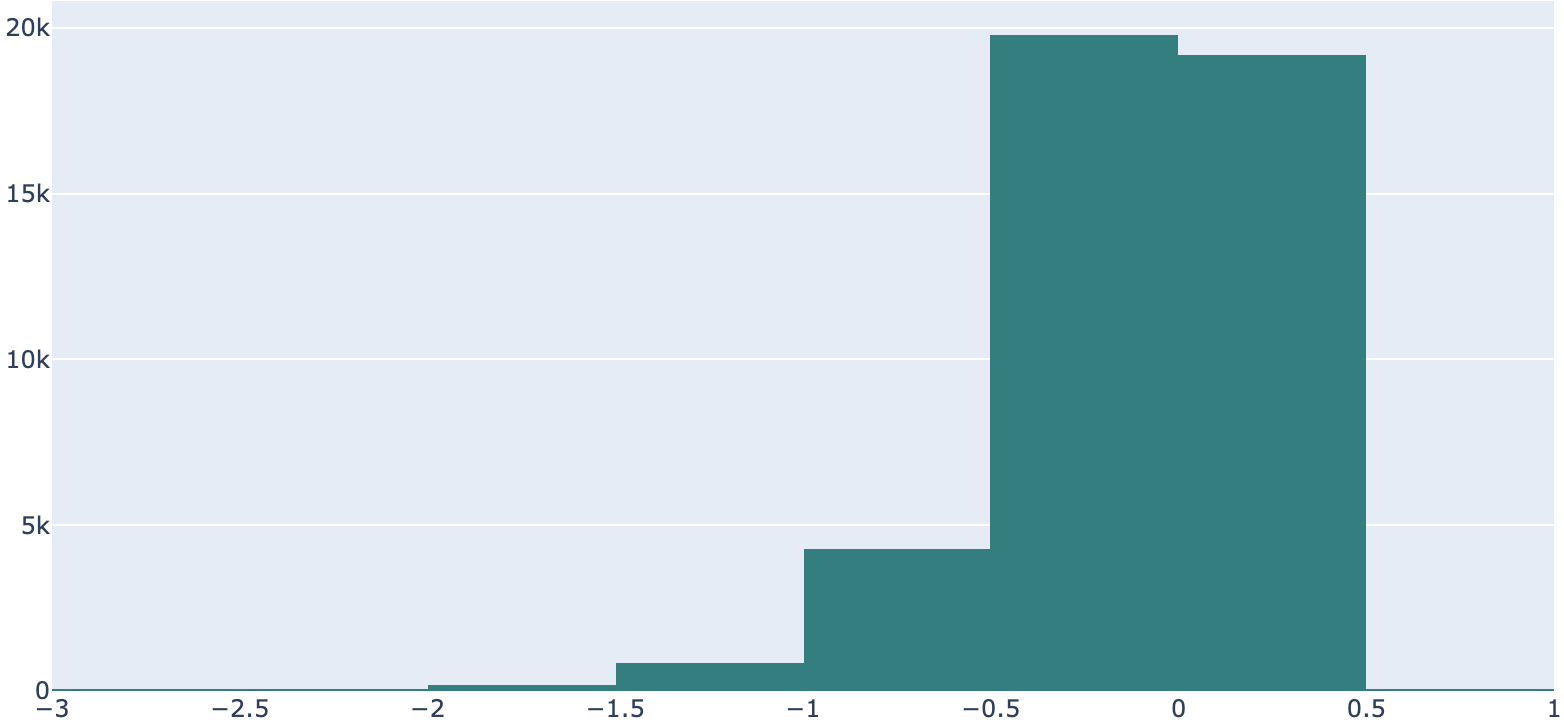}} \;\;\;
     \subfloat[]{\includegraphics[width =0.3\textwidth, height = 0.18\textwidth]{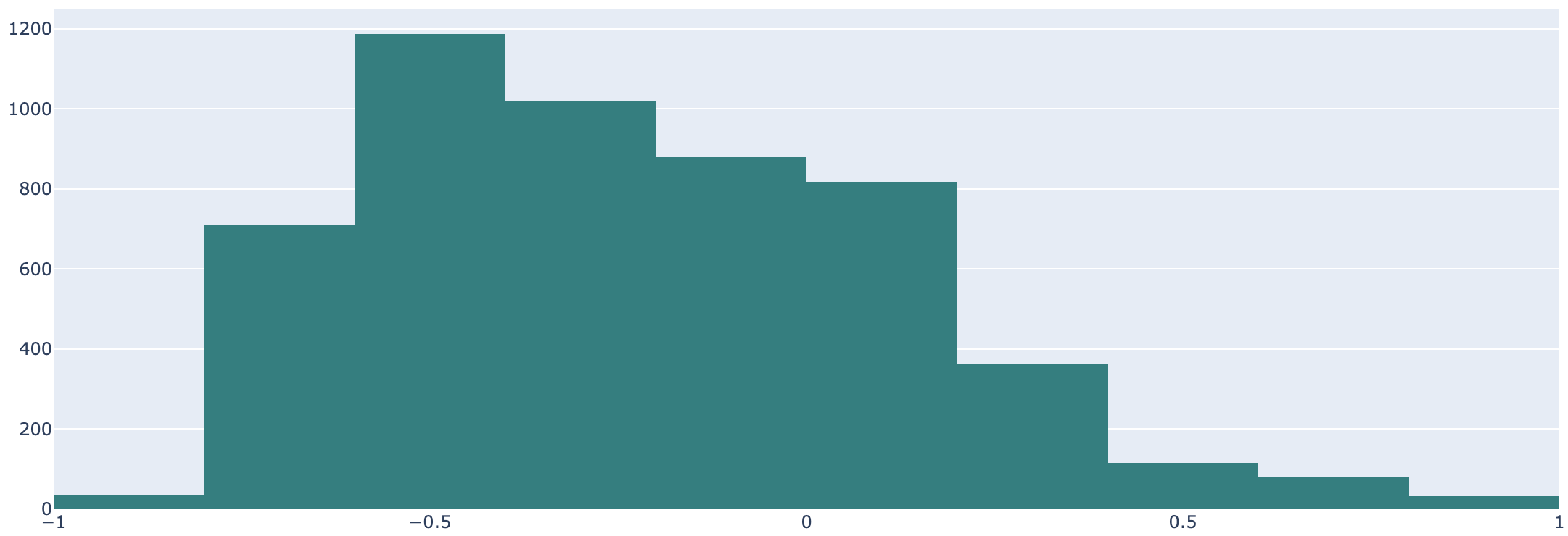}}\;\;\;
     \subfloat[]{\includegraphics[width =0.3\textwidth, height = 0.18\textwidth]{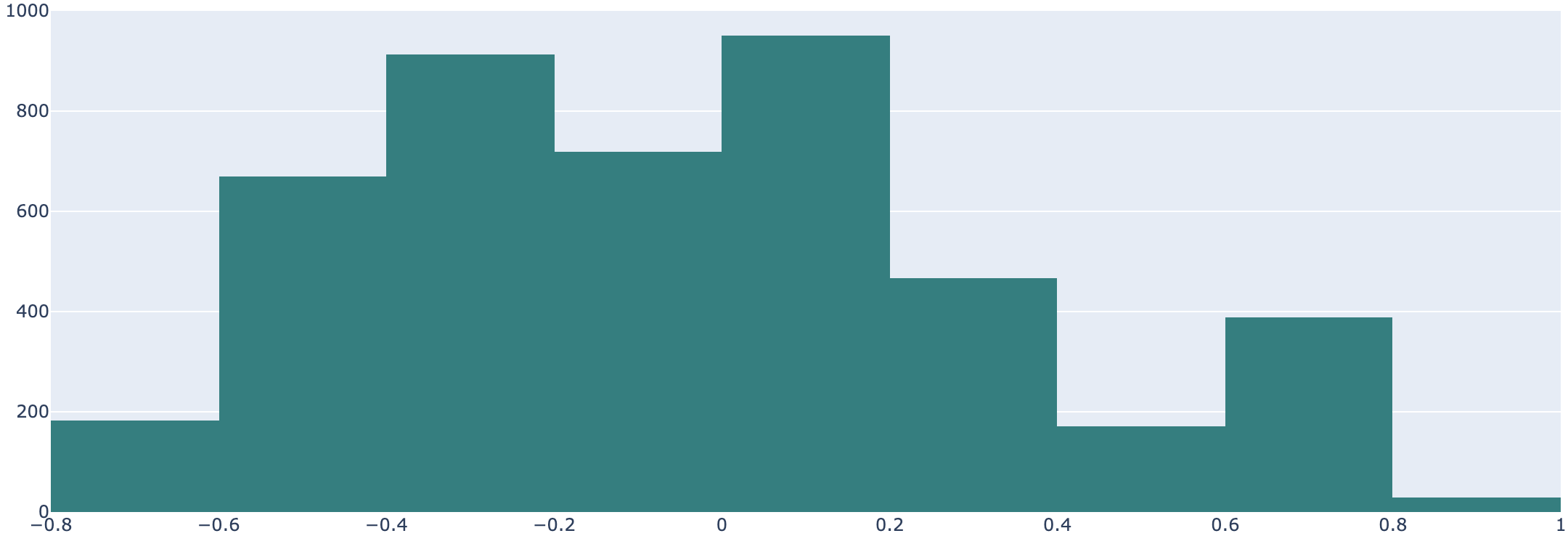}}\;\;\;
     \subfloat[]{\includegraphics[width = 0.3\textwidth, height = 0.18\textwidth]{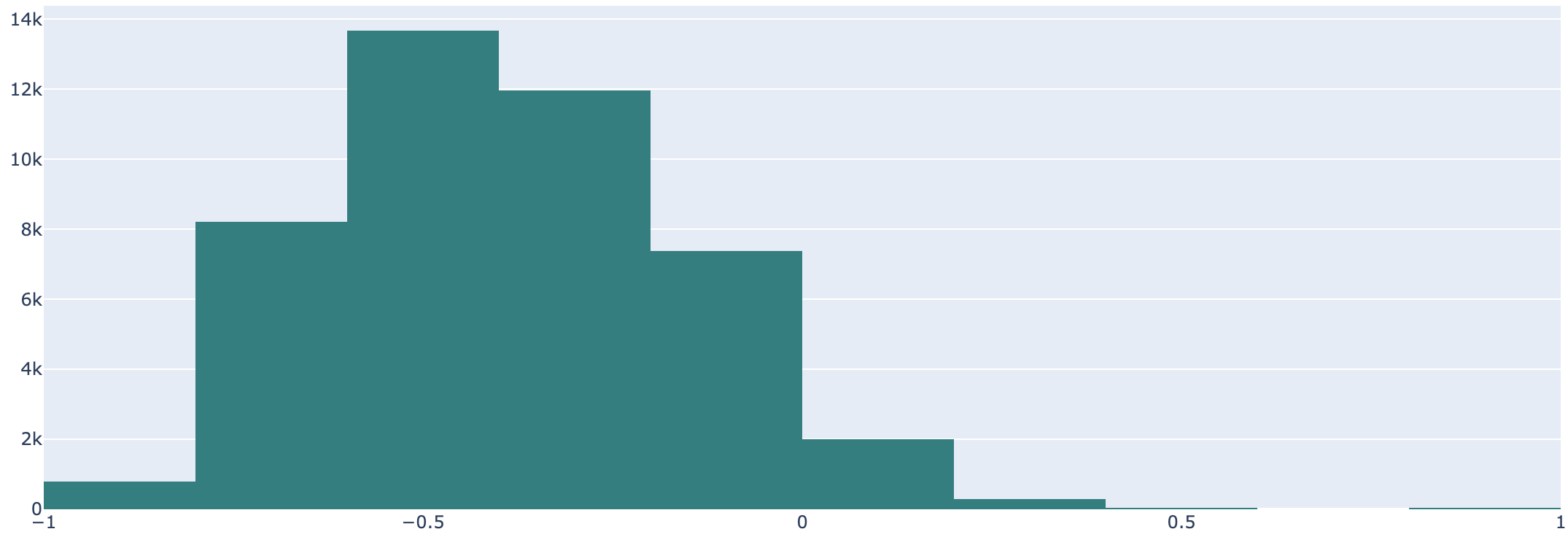}} \;\;\;
\caption{Changes of the graph Ricci curvature distribution before (first row) and after (second row) the computation conducted in RC-UFG (Hom). The datasets on each row from left to right are \textbf{Cora, Citeseer} and \textbf{Pubmed}, we fixed $\alpha =0.4$ as the initial mass for curvature computation. We observe the $\zeta^\text{Hom}(\kappa_{i,j})$ is able to smooth the curvature of graph by contracting both the negative and positive curvatures.}  
\label{ricci_flow_fig}
\end{figure}
It is easy to check that based on the range of graph Ricci curvature, all weights generated from $\zeta^\text{Hom}$ are positive and $\zeta^\text{Hom}$ is a bijective function thus the form of $\zeta^\text{Hom}$ satisfies the basic conditions. Furthermore, it has been studied in \cite{ni2019community} that graph Ricci flow smooths both positive and negative curvatures towards some specific quantities (mostly 0), and Fig.~\ref{ricci_flow_fig} shows this smoothing phenomenon of $\zeta^\text{Hom}$ on citation networks. We now provide some analysis on how this smoothing effect introduced from $\zeta^\text{Hom}$ can help framelet adapt better in homophily graph. For convenience, in the sequel, we will call the framelet that equips with curvature information as RC-UFG (Hom) which contains the (curvature enhanced) graph Laplacian  (denoted as $\widehat{\mathbf L}_\kappa$) as 
\begin{equation}\label{e:RCLap}
    \widehat{\mathbf L}_\kappa = \mathbf I- \widetilde{\mathbf D}_\kappa ( \mathbf A \odot \boldsymbol{\zeta^\text{Hom}_\kappa}  + \mathbf I) \widetilde{\mathbf D}_\kappa,
\end{equation}
where $\widetilde{\mathbf D}_\kappa$ is the node degree obtained from the sum of each row of $\mathbf A \odot \boldsymbol{\zeta^\text{Hom}_\kappa} + \mathbf I$, and $\boldsymbol{\zeta^\text{Hom}_\kappa} \in \mathbb R^{n\times n}$ is the matrix with entries as $\zeta^\text{Hom}(\kappa_{i,j})$. We first conclude that without curvature information inserted, framelet convolution can induce both LFD and HFD in the following lemmas: 
\begin{lem}
\label{framelet_LFD_HFD_initial}
The spectral graph framelet convolution \eqref{spectral_framelet} with Haar-type filter can induce both LFD and HFD dynamics. Specifically, let $\mathbf \theta_{0,J} = \mathbf 1_n$ and $\mathbf \theta_{r,j} = \theta \mathbf 1_n$ for $r=1,...,L, j = 1,...,J$ where $\mathbf 1_n$ is a vector of all $1$s. Suppose $\theta \geq 0$. Then when $\theta \in [0,1)$, the spectral framelet convolution is LFD and when $\theta > 1$, the spectral framelet convolution is HFD.
\end{lem}
The proof of the lemma relies on \cite{han2022generalized} in which similar conclusion is shown mainly for spatial framelet, here based on the idea in \cite{han2022generalized} we show a stretch of proof for spectral framelet. The core idea of the proof is to analyze the gradient flow of the total Dirichlet energy (please refer to \cite{han2022generalized} for more details) of graph framelet, and such gradient flow (with a stepsize $\tau$) has the form of: 
\begin{align}\label{eq_vec_prop}
    &\vec \big( \mathbf H(m\tau) \big) = 
    \tau^m \Big( \mathbf W \otimes (\mathcal W_{0,1}^\top \mathcal W_{0,1} + 
    \theta \mathcal W_{1,1}^\top \mathcal W_{1,1}) \Big)^m \! \vec \big( \mathbf H(0) \big) \notag \\
    &= \tau^m \sum_{k,i} \Big(\lambda_k^W  \big(  \cos^2(\lambda_i/8) + \theta \sin^2(\lambda_i/8) \big) \Big)^m \!\!\!c_{k,i}(0) \mathbf \phi_k^W \otimes \mathbf u_i.
\end{align}
where $\{ (\lambda_k^W, \mathbf \phi_k^W) \}_{k = 1}^c$ denotes the eigenvalue and eigenvector pairs of $\mathbf W$ and $c_{k,i}(0) := \langle  \vec(\mathbf H(0)), \mathbf  \phi_k^W \otimes \mathbf u_i \rangle$. Now we see
\begin{align*}
    |\lambda_k^W (\cos^2(\lambda_i/8) + & \theta \sin^2(\lambda_i/8) )| 
    \leq \Delta^W (\cos^2(\lambda_i/8) + \theta \sin^2(\lambda_i/8) ).
\end{align*}
where $\Delta^W := \max_k |\lambda_k^W|$. One can easily check when $\theta > 1$, $\cos^2(\lambda_i/8) + \theta \sin^2(\lambda_i/8)$ is (monotonically)  increasing in $\lambda_i \in [0, \rho_{\widehat{\mathbf L}}]$ and its maximum is achieved at $\lambda_i = \rho_{\widehat{\mathbf L}}$ (with sufficiently large $\lambda_k^W$), this suggests the dominant frequency is $\rho_{\widehat{\mathbf L}}$, and thus a HFD dynamic. On the other hand, when $\theta \in [0,1)$, the function is (monotonically) decreasing and the maximum is achieved at $\lambda_i = 0$. With the same reasoning, we see the model is LFD, regardless of $\lambda_k^W$. More precisely, let $\delta := \max_{i: \lambda_i \neq \rho_L} \vert \big( \lambda^{W}(  (\lambda_i^{\Lambda_{0,1}})^2 +  \theta (\lambda_{i}^{\Lambda_{1,1}})^2) \big)  \vert$. Also denote $\mathbf P_\rho = \sum_{k} (\boldsymbol{\phi}_{k} \otimes \mathbf u_\rho)(\boldsymbol{\phi}_{k} \otimes \mathbf u_\rho)^\top$ where $\mathbf u_\rho$ is the eigenvector of $\widehat{\mathbf L}$ associated with eigenvalue $\rho_{\widehat{\mathbf L}}$ (assuming the eigenvalue $\rho_{\widehat{\mathbf L}}$ is simple). Then we can decompose Eq.~\eqref{eq_vec_prop} as 
\begin{align*}
    &\vec\big( \mathbf H(m\tau) \big) = \tau^m \sum_{k } \delta_{\rm HFD}^m  c_{k, \rho_L}(0) \mathbf u_k \otimes \mathbf u_\rho  \!\! +\!\! \tau^m \sum_{k} \sum_{i : \lambda_i \neq \rho_L} \Big( \big( \lambda^{W} (\lambda_{i}^{\Lambda_{0,1}})^2  
    +  \theta(\lambda_{i}^{\Lambda_{1,1}})^2 \big)  \Big)^m c_{k, i}(0) \boldsymbol{\phi}_k \otimes \mathbf u_\rho \\
    &\leq \tau^m \delta_{\rm HFD}^m (  \mathbf P_\rho \vec \big( \mathbf H(0) \big) + \!\!\!
    \sum_{k} \!\!\! \sum_{i : \lambda_i \neq \rho_L} \!\!\! \left( \frac{ \delta }{\delta_{\rm HFD}} \right)^m c_{k, i}(0) \boldsymbol{\phi}_k \otimes \mathbf u_\rho ,
\end{align*}
 where $\delta < \delta_{\rm HFD}$.
By normalizing the results, we obtain $\frac{\vec\big( \mathbf H(m\tau) \big)}{\|  \vec\big( \mathbf H(m\tau) \big)\|} \xrightarrow{} \frac{\mathbf P_\rho (\vec(\mathbf H(0)))}{\| \mathbf P_\rho \vec (\mathbf H(0)) \|}$, as $m \xrightarrow{} \infty$, where the latter is a unit vector $\mathbf h_{\infty}$ satisfying $(\mathbf I_c \otimes \widehat{\mathbf L}) \mathbf h_{\infty} = \rho_L \mathbf h_{\infty}$. This suggests, the dynamics is HFD according to Definition \ref{def_hfd_lfd} and Lemma \ref{lemma_hfd_lfd}.

\begin{rem}\cite{han2022generalized}
The conclusion (i.e., HFD and LFD induction) in Lemma \ref{framelet_LFD_HFD_initial} also holds for spatial framelet.    
\end{rem}
Different from \cite{han2022generalized} in which the control of the scaling diagonal matrix $\text{diag}(\theta)$ is applied to framelet to induce LFD/HFD, we show that with the help of $\zeta$, our curvature re-weighting scheme can also achieve the same goal, and this is based on the impact of $\zeta$ on the spectrum of the graph (i.e., the eigenvalue of graph Laplacian) in the following lemma: 
\begin{lem}[Theorem 4 in \cite{bauer2011ollivier}]
\label{lap_eigen_range}
For any finite graph $\mathcal G(\mathcal V_\mathcal G, \mathcal E_\mathcal G)$, if $\kappa_{i,j} > k$, then $\rho_{\widehat{L}} \leq 2-k$ and $\lambda_1 \geq k$.
\end{lem}
Obviously $k$ can be $\min\{\kappa_{i,j}\}$. As we have illustrated previously, we can treat $\zeta^\text{Hom}(\kappa_{i,j}) = d_s{(i,j)} (1 - \kappa_{i,j}) = 1 -\kappa_{i,j}$ as the first step of graph Ricci flow which shrinks both positive and negative curvatures towards 0, therefore we have the smallest curvature (assuming the input graph contains negative curvature) increase ($k$ increase) and largest curvature decrease. Hence we have $\rho_{\widehat{L}}$ decrease according to Lemma \ref{lap_eigen_range}. Consequentially, when framelet is LFD (i.e., $\theta <1$), the insertion of curvature information further restricts the range of graph Laplacian and achieving a lower frequency dominant than the previous LFD framelet, and thus a better adaption to homophily graph dataset. 


\subsection{Conditions on $\zeta$ when the input graph is heterophilic}\label{heterophily_adaption}

\subsubsection{Form of $\zeta^\text{Het}$}

As we illustrated before, when the input graph is heterophilic, a sharpening effect of GNN is preferred. To achieve this goal, a further investigation between edge weights and Ricci curvature is needed, this is summarized in the following lemma:
\begin{lem}[Theorem 4 in \cite{jost2014ollivier}]
\label{lower_bound}
On a weighted locally finite graph $\mathcal G(\mathcal V_\mathcal G, \mathcal E_\mathcal G)$ we have:
\begin{align*}
    \kappa_{i,j} &\geq -2\left(1-\frac{a_{i,j}}{d_i} - \frac{a_{i,j}}{d_j}\right)_+ 
=\begin{cases}
    -2+ \frac{2a_{i,j}}{d_i}+\frac{2a_{i,j}}{d_j},\ \text{if } d_i, d_j >1 \\
    0,\ \text{Otherwise}
\end{cases}    
\end{align*} 
where $s_+ = \max\{0,s\}$.
\end{lem}
The weighted trees attain this lower bound. Based on Lemma \ref{lap_eigen_range}, the largest eigenvalue of graph Laplacian is bound by $2-k$ where $k$ is the smallest Ricci curvature. Together with Lemma \ref{lower_bound} above, to enforce the heterophily adaption, we shall ensure the edge weights generated from $\zeta^{\text{Het}}(\widetilde{\kappa}_{i,j})$ are lesser than the current edge weights, ideally for all edges.
Furthermore, as the graph is initially unweighted, we have all $a_{i,j}  = 1$, thus an ideal function $\zeta^\text{Het}$ shall be a bijective function that maps all input curvatures to the range of $[0,1]$. To do this, we first normalize the graph Ricci curvature into a symmetric range of $[-1,+1]$ and the denote the normalized Ricci curvatures as $\widetilde{\kappa} $. Then the form of $\zeta^\text{Het}(\widetilde \kappa)$ is: 
\begin{align}\label{zeta_hetero}
    \zeta^\text{Het}(\widetilde \kappa_{i,j}) = \frac{1+\widetilde \kappa_{i,j}}{2}.
\end{align}
For the rest of the paper, we start to call the framelet model with the curvature information inserted from $\zeta^\text{Het}(\widetilde \kappa)$ as RC-UFG (Het). Thus geometrically one can interpret $\zeta^\text{Het}(\widetilde \kappa)$  as a scaled normalized Ricci flow in the opposite direction (i.e., an edge metric deformation that sparse the graph rather than smooths it out). Based on Eq ~\eqref{zeta_hetero}, all the induced edge weights are less than the initial edge weights and thus resulting a smaller lower bound of the graph Ricci curvature. According to Lemma \ref{lap_eigen_range} we have the largest eigenvalue of graph Laplacian increase and this leads a stronger energy convergence (HFD) based on Lemma \ref{framelet_LFD_HFD_initial}.

\subsubsection{Curvature based edge dropping}\label{sec:CBED}
Graph edge dropping method is originally introduced in \cite{rong2019dropedge} in which a random edge dropping scheme is designed for alleviating over-fitting and over-smoothing problem in GNN. In this section, we show that we can drop graph edges based on graph curvature information and thus make framelet model produce more sharpening effect than smoothing to some extend. To see this, we first show some deeper links between graph Ricci curvature and the topology of the graph. Specifically, recall that geometrically, Ricci curvature describes whether two geodesics from diverging (resp. converging) away. In such situation, two geodesics starting from different directions and eventually converging to a point, would form a triangular structure. As the graph Ricci curvature shows the difficulty of information transferring from one node along one edge to another, 
if there are a large number of triangles overlaps between two nodes, information will be easier to pass through, resulting in a lower optimal transportation distance and a larger Ricci curvature. This observation is aligned with the so-called graph local clustering coefficient \cite{chung1997spectral} which is defined as: 
\begin{align}
    c(i) = \frac{1}{d_i(d_i-1)}\sum_{j, j\in \mathcal N_i} \#(i,j),
\end{align}
where $\#(i,j)$ is the number of triangles which includes the node $i$ and $j$ as its vertices. It is easy to verify that if the graph is fully connected (i.e. complete graph), $c(i) =1$ for all $i$, if the graph nodes are almost totally isolated with each other, then $c(i)$ is approaching to 0. Therefore one may consider to reduce the overlapped triangles between two nodes to make the graph locally sparser. In fact, based on \cite{jost2014ollivier} we have the following inequality holds for all edges: 
\begin{align}\label{triangle_curvature}
    \kappa_{i,j} \leq \frac{\#(i,j)}{d_i \vee d_j },
\end{align}
where ${d_i \vee d_j } = \max\{d_i,d_j\}$. It is obvious that pruning graph edges leads to a modification on the graph Ricci curvature. However, the next question is, \textit{what edges are preferable to drop and keep?}

To address this problem, we first consider the functionality of Ricci curvature via different signs. We note that to explicitly show how curvature with different signs can affect framelet learning outcome, in the following analysis, we take spatial framelet as an example although the relationship between spatial and spectral framelet has been studied in \cite{chendirichlet}. Recall that spatial framelet is with an adjacency information based propagation rule as: $\mathbf H(\ell + 1) =\sigma\left(\sum_{r,j \in \mathcal I }   \mathcal W_{r,j}^\top \widehat{\mathbf A}\mathcal W_{r,j} \mathbf H(\ell) \mathbf W^\ell_{r,J}  \right)$, where we denote the index set $\mathcal I = \{(r,j) : r = 1,...,L, j = 1,...,J \} \cup \{ (0, J) \}$. One can easily check that each layer of spatial framelet aggregates the node neighbouring information under the domain specified by $\mathcal W_{r,J}$, and thus can be treated as a type of message-passing neural network (MPNN) \cite{gilmer2017neural}with the form as: 
\begin{align}\label{mpnn}
    \mathbf h^{(\ell+1)}_i = \phi_\ell \left(\bigoplus_{u\in \mathcal N_i} (\mathcal W_{r,J})\psi_\ell( \mathbf h^{(\ell)}_i, \mathbf h^{(\ell)}_u) \right),
\end{align}
where $\mathcal N_i$ stands for the set of all neighbours of node $i$, $\phi_\ell$ is the updated function, usually presented as the activation function, $\bigoplus$ is the aggregation function and $\psi_\ell$ is the message passing function which is usually trainable. We now show that it is those very positively-curved edges (i.e.,$\kappa_{i,j} \approx 1$) that makes the node features become similar (and potentially indistinguishable) when the number of layer becomes higher, and thus shall be the target for the edge dropping. 

Without loss of generality, we let $d_i > d_j$, and we quantify the difference between two node ($\mathbf h_i$ and $\mathbf h_j$) features generated from spatial framelet as layer $\ell+1$ in terms of message-passing form as: 
\begin{align}\label{edge_drop_derive}
    &\left|\mathbf h^{\ell+1}_i - \mathbf h_j^{\ell+1}\right| = \left|\sigma_\ell\left(\sum_{(r,j) \in \mathcal I }   \mathcal W_{r,j}^\top \widehat{\mathbf A} \mathcal W_{r,j} \mathbf H(\ell) \mathbf W^\ell_{r,J}  \right)_i 
    - \sigma_\ell\left(\sum_{(r,j) \in \mathcal I }   \mathcal W_{r,j}^\top \widehat{\mathbf A} \mathcal W_{r,j} \mathbf H(\ell) \mathbf W^\ell_{r,J}  \right)_j \right | \notag \\
    &=\left |\phi_\ell \left(\bigoplus_{u\in \mathcal N_i} (\mathcal W_{r,J})\psi_\ell( \mathbf h^{(\ell)}_i, \mathbf h^{(\ell)}_u) \right)-\phi_\ell \left(\bigoplus_{v\in \mathcal N_j} (\mathcal W_{r,J})\psi_\ell( \mathbf h^{(\ell)}_j, \mathbf h^{(\ell)}_v) \right)\right| \notag\\
    &\leq \mathcal S\left|\left(\bigoplus_{u\in \widetilde{\mathcal N}_i} (\mathcal W_{r,J})\psi_\ell( \mathbf h^{(\ell)}_i, \mathbf h^{(\ell)}_u) \right)- \left(\bigoplus_{v\in \widetilde{\mathcal N}_j} (\mathcal W_{r,J})\psi_\ell( \mathbf h^{(\ell)}_j, \mathbf h^{(\ell)}_v) \right)\right|,  
\end{align}
where the last inequality is obtained by assuming $\phi_\ell$ is $\mathcal S$ Lipschitz, and $\widetilde{\mathcal N}_i$,  $\widetilde{\mathcal N}_j$ are the set of neighbours for node $i$ and $j$ including themselves. Since we have assumed $d_i > d_j$ thus the extra neighbouring information for node $i$ compared to node $j$ is $|\widetilde{\mathcal N}_i \setminus \widetilde{\mathcal N}_j| = n+1 - \#(i,j)-2$. Plugging in this into Eq.~\eqref{triangle_curvature} we have $|\widetilde{\mathcal N}_i \setminus \widetilde{\mathcal N}_j| \leq n-n\kappa_{i,j} = n(1-\kappa_{i,j})$. This result immediately suggests that difference between feature representation tend to become 0 if the edge is very positive (i.e., closed to 1)\footnote{Precisely speaking to obtained this conclusion, we shall further assume that the message passing function is bounded i.e., $|\phi_\ell (\mathbf h_i) |\leq \mathbf M |(\mathbf h_i)|, \quad \forall \mathbf h \in \mathbf H$}. Therefore, those edges with very positive curvatures are the targets one shall focus on for edge dropping. We summarize the steps of curvature based edge dropping (CBED) in Algorithm \ref{Alg1}. Finally, one can easily check that once framelet model is HFD, dropping edges to the input graph will induce a stronger HFD, and this observation aligns with the conclusions from recent studies \cite{giraldo2022understanding}.


\begin{algorithm}[t]
\caption{Curvature based graph edge dropping (CBED) } \label{Alg1}
\begin{algorithmic}[1] 
\STATE \textbf{Input:} graph $\mathcal G$, maximum iteration number K, $\kappa$ upper bound, number of cutting $\epsilon \in \mathbb Z^+$ in each iteration, target $\kappa$ upper bound
\REPEAT  
\FOR{edge $e_{i,j} \in \mathcal E_\mathcal G$ with maximal  $\kappa$}
    \STATE{randomly remove the edge  that belongs to the triangular structure that contains $e_{i,j}$ as its edge.}\ENDFOR
\UNTIL{target $\kappa$ upper bound or maximum iteration number reached}
\end{algorithmic}
\end{algorithm}

\begin{rem}[Compared to graph rewiring in \cite{topping2021understanding}]
The recent work \cite{topping2021understanding} developed a curvature based graph rewiring process to alleviate the so-called over-squashing problem. The rewiring method in \cite{topping2021understanding} targets on those edges with negative (balance Forman) curvatures and by building additional tunnel (connectivity) to support those edges, in the meanwhile, drop the edges with the most positive curvature, so that GNN is thus easier to aggregate the multi-hop neighbouring information which tends to dilute when the number of layer of GNNs becomes higher. However, in CBED, we focus on the changes of framelet asymptotic behavior when model's smoothing effect is diluted.    
\end{rem}

\begin{rem}[CBED as hard verison of $\zeta^\text{Het}$]\label{soft_hard}
Based on the algorithm one can interpret the RC-UFG (Het) with its curvature information inserted from $\zeta^\text{Het}(\widetilde \kappa)$ as a soft version of CBED as the graph topology (connectivity) is preserved in RC-UFG (Het) whereas CBED cuts the unnecessary edges (for heterophily adaptation purpose) and directly nullifies  weights. In practice, when the input graph is dense, the re-weighting effect induced from RC-UFG (Het) may not be strong enough as we may have $d_i \gg w_{i,j}$. In this case, CBED is obviously preferred as it quickly reduces the edge weight.
\end{rem}

\section{Experiment} \label{experiment}

In this section, we show a variety of numerical tests for our curvature enhanced graph convolution. Section \ref{node_prediction} shows the performance of RC-UFG (Hom) on node classification on the real-world homophily graph datasets. Then in Section \ref{heterophily_section} we present the testing outcomes of RC-UFG (Het) as well as RC-UFG (Het) plus CBED models on heterophily graph datasets. In addition, we conduct ablation study in Section \ref{sec:ablation} by inserting the curvature information from $\zeta^\text{Hom}$ or $\zeta^\text{Het}$ to the baseline models to show the effectiveness of incorporating the curvature based methods to other GNNs. Lastly, in Section \ref{sec:computation_complexity} we include a discussion on the computational complexity of graph Ricci curvature and how such computation can be approximated by different optimal transport solvers.
All experiments were conducted using PyTorch on NVIDIA\textsuperscript{\textregistered} Tesla V100 GPU with 5,120 CUDA cores and 16GB HBM2 mounted on an HPC cluster. The source code of the paper can be found in \url{https://github.com/dshi3553usyd/curvature_enhanced_graph_convolution}.




\subsection{RC-UFG (Hom) For Node Classification } \label{node_prediction}

\paragraph{Dataset} We tested RC-UFG (Hom) model against the state-of-the-arts on five node classification datasets.
The first task for node classification is conducted on several benchmark citation networks. The graph datasets \textbf{Cora, Citeseer} are relatively small and sparse with average node degree below 2. Other datasets, \textbf{Coauthor CS} is the co-authorship graphs based on the Microsoft Academic Graph from the KDD Cup 2016 challenge;  \textbf{Amazon Photos} is the segment of the Amazon co-purchase graph. These datasets together with \textbf{PubMed} have more than 10 thousands nodes and 20 thousands edges and with average nodes degrees more than 20, hence are denser and larger than \textbf{Cora} and \textbf{Citeseer}. Table \ref{datasets_statistics} shows some basic statistics of the citation datasets. 

\begin{table}[t]
\centering
\caption{Statistics of the homophily benchmarks, $H(G)$ represent the level of homophily of overall benchmark datasets. }
    \label{datasets_statistics}
\setlength{\tabcolsep}{2pt}
\renewcommand{\arraystretch}{1.8}
    \begin{tabular}{ccccccc}
    \hline
         \textbf{Datasets} & \#Class & \#Feature & \#Node & \#Edge & Train/Valid/Test & H(G)\\
         \hline
        \textbf{Cora} & 7 & 1433 & 2708 & 5278 & 20\%/10\%/70\% & 0.825\\
        \textbf{CiteSeer} & 6 & 3703 & 3327 & 4552 & 20\%/10\%/70\%  & 0.717\\
        \textbf{PubMed} & 3 & 500 & 19717 & 44324 & 20\%/10\%/70\%  & 0.792\\
        \textbf{Photo} & 8 & 745 & 7487 & 119043 & 20\%/10\%/70\%  & 0.849\\
        \textbf{CS} & 15 & 6805 & 18333 & 81894 & 20\%/10\%/70\%  & 0.832\\
        \hline
\end{tabular}
\end{table}

\paragraph{Set up}
The RC-UFG (Hom) is designed  with two curvature information based convolution layers for learning the graph embedding the output, which is followed by softmax activation function for the final prediction. Most of hyperparameters are with the default values except from learning rate, weight decay, hidden units, dropout ratio and curvature initial weight index $\alpha$ in training. The grid search  was applied to fine tune  these hyperparameters. The search space for learning rate is in $\{0.1, 0.05, 0.01, 0.005\}$, number of hidden units in $\{16, 32, 64\}$,and weight decay in $\{0.05, 0.01,0.015,0.2\}$.
We set the maximum number of epochs of 200 for citation networks. All the datasets included in this series of experiment are split followed by the standard public processing rules. All the average test accuracy and standard deviations are summarized from 10 random trials.

\paragraph{Baseline} 
We consider eight baseline models (listed below) for comparison. In addition, to show the benefits of choosing $\zeta^\text{Hom}$, we also included RC-UFG (Het) in this experiment. 
\begin{itemize}
\item \textbf{MLP}: The standard full-connected feed-ward multiple layer perceptron. 
 
\item \textbf{GCN}: First developed in \cite{kipf2016semi}, GCN is the first kind of linear approximation to the spectral graph convolutions.

\item \textbf{MoNet}: MoNet \cite{monti2017geometric} proposes a unified framework, allowing to generalize classic CNN architectures to non-Euclidean domains (graphs and manifolds) and learn
local, stationary, and compositional task-specific features.

\item \textbf{GAT}: GAT is the model that first assigns the attention mechanism to the graph structured data.  

\item \textbf{JKNet}: JKNet \cite{xu2018representation} flexibly leverages neighbouring information via different ranges (jumping knowledge) and thus propagating the node information from  deeper graph structure.

\item \textbf{APPNP} : APPNP \cite{DBLP:conf/iclr/KlicperaBG19} merges the GNN frame with the famous personal PageRank to separate the neural network from the propagation scheme.

\item \textbf{GPRGNN} 
As a generalized version of PageRank in GNN, GPRGNN \cite{chien2020adaptive}is capable of producing a learnable weight score to the GNN output by capturing both node feature and graph topological information. GPRGNN is proved to be always escape from over-smoothing issue. 

\item \textbf{CurvGN} The CurvGN \cite{ye2019curvature} is the work that firstly (to our best knowledge) assign curvature information to the GNN model.  

\item \textbf{UFGConv} \cite{zheng2021framelets}: UFG is a type of GNNs based on framelet transforms, the framelet decomposition can naturally aggregate the graph features into low-pass and high-pass spectra. In addition, we included two variants of framelet: UFGConv\_S (shrinkage) and UFGConv\_R (Relu).
\end{itemize}


\begin{table}[t]
\centering
\setlength{\tabcolsep}{3.3pt} 
\renewcommand{\arraystretch}{1.6} 
\caption{Test Accuracy (in percentage) for citation networks with standard deviation after $\pm$. The top  results is highlighted in \textbf{bold}.}
\label{table1}
\begin{tabular}{llllll}
\hline
\textbf{Method} & \textbf{Cora} & \textbf{Citeseer} & \textbf{Pubmed} & \textbf{CS}     & \textbf{Photo} \\ \hline
MLP             & 55.1          & 59.1              & 71.4            & $88.3{\pm} 0.7$ & $69.6{\pm}3.8$ \\
MoNet           & 81.7             & 71.2              & 78.6              & $90.8 {\pm} 0.6$     & $91.2 {\pm}1.3$       \\
GCN            & $81.5 {\pm} 0.5$ & $70.9 {\pm} 0.5$  & $79.0 {\pm} 0.3$  & $91.1 {\pm} 0.5$     & $91.2 {\pm} 1.2$      \\
GAT             & $83.0 {\pm} 0.7$ & 72.5\(\pm\)0.7  & $79.0 {\pm} 0.3$  & $90.5 {\pm} 0.6$     & $85.1 {\pm} 2.3$      \\
JKNet           & $83.7{\pm}0.7$             &     $72.5{\pm}0.4$           &  $82.6{\pm}0.5  $             &       $91.1{\pm}0.3  $        &       $86.1{\pm}1.1$         \\
APPNP           &   83.5\(\pm\)0.7           &  \textbf{75.9\(\pm\)0.6}                &    $80.2{\pm}0.3 $        &   $91.5{\pm}0.1 $             &  $87.0{\pm}0.9 $             \\
GPRGNN          & $83.8{\pm}0.9$              &  $75.9{\pm}0.7$                 &   $82.3{\pm}0.2$              &  $91.8{\pm}0.1$               &    $87.0{\pm}0.9$            \\
CurvGN          & $82.6 {\pm} 0.6$ & $71.5 {\pm} 0.8$  & $78.8 {\pm} 0.6$  & $92.9  {\pm} 0.4$    & $92.5 {\pm}0.5$       \\
UFGConv\_S      & $83.0 {\pm} 0.5$ & $71.0 {\pm} 0.6$  & $79.4 {\pm} 0.4$  & $92.1 {\pm} 0.2$     & $92.1 {\pm} 0.5$       \\ 
UFGConv\_R      & 83.6\(\pm\)0.6 & $72.7\pm0.6$  & 79.6\(\pm\)0.4 & 93.0\(\pm\)0.7     & 92.5\(\pm\)0.2 \\ \hline
RC-UFG (Hom)     & \textbf{84.4\(\pm\)0.7} & 72.5\(\pm\)0.7  & \textbf{82.9\(\pm\)0.2} & \textbf{94.2\(\pm\)0.9}    & \textbf{93.5\(\pm\)0.7} \\

RC-UFG (Het) & 80.6\(\pm\)0.4 & 71.7\(\pm\)0.6  & 79.6\(\pm\)0.4 & 90.4\(\pm\)1.2     & 89.5\(\pm\)1.9 \\ \hline
\end{tabular}
\end{table}

\paragraph{Results}
The best test accuracy score (in percentage) are highlighted in Table \ref{table1}.RC-UFG (Hom) model achieves the highest prediction in most of benchmark datasets compared to baseline models. In particular, RC-UFG (Hom) tends to perform better when the graph is more homophilic (i.e., the result in \textbf{CS} and \textbf{Photo}), indicating the theoretical correctness of the smoothing effect from graph Ricci flow (i.e., $\zeta^\text{Hom}$ on the edge weights). Furthermore, the perform of RC-UFG (Hom) is much better than RC-UFG (Het), also suggesting a stronger smoothing effect is preferred when the graph is homophilic.

\subsection{Node Classification in Heterophily Graph}
\label{heterophily_section}
In this section, we show the performance of RC-UFG (Het) on heterophily graph datasets. 

\paragraph{Data sets and Baselines}
We compare the learning outcomes of RC-UFG (Het) to various baseline models, for the experiment on heterophily graph datasets, we also include the following additional baselines as follows: 
\begin{itemize}
\item \textbf{H2GCN}: H2GCN is combines both low and high order neighbouring information and intermediate node representation to boost the learning outcome when the input graph is heterophily. 
 
\item \textbf{Mixhop}: Mixhop model \cite{abu2019mixhop} can capture graph neighbouring relationships by repeatedly mixing feature representations of neighbors at various distances, in the meanwhile, requires no additional memory or computational complexity.

\item\textbf{GraphSAGE}: GraphSAGE \cite{hamilton2017inductive} is a general inductive framework which learns a function that generates embeddings by sampling and aggregating features
from a node’s local neighborhood.
\end{itemize}
We test the mentioned baseline models for 10 times on \textbf{Cornell, Wisconsin, Texas, Film, Chameleon and Squirrel} following the same early stopping strategy. Table \ref{hetero_datasets_statistics} presents the summary statistics of the included heterophily graphs. Moreover, we also include the maximal and minimal graph Ricci curvature of included datasets. Similar to what we designed for homophily graph, we also include RC-UFG (Hom) to make a fair comparison. Furthermore, we added one additional RC-UFG (Het) model with CBED which we set the upper Ricci curvature after CBED as 0.7 for all datasets, and we denote the RC-UFG (Het) with CBED as RC-UFG (Het)\_D.

\begin{table}[t]
\caption{Statistics of the homophily benchmarks, $H(G)$ represent the level of homophily of overall benchmark datasets. All dataset are spilt with 60\% for training, and 20\% for testing and validation.}
\label{hetero_datasets_statistics}
\centering
\setlength{\tabcolsep}{3.5pt}
\renewcommand{\arraystretch}{2}
    \begin{tabular}{ccccccc}
    \hline
         \textbf{Datasets} & \#Class & \#Feature & \#Node & \#Edge & Max,Min($\kappa$) & H(G)\\
         \hline
        \textbf{Chameleon} &5 &2325 &2277 &31371 &0.91,-1.16 &0.247\\
        \textbf{Squirrel} &5 &2089 &5201 &198353 &0.78,-1.12 &0.216\\
        \textbf{Actor} &5 &932 &7600 &26659 & 0.85, -1.17
        
        &0.221\\
        \textbf{Wisconsin} &5 &251 &499 &1703 &0.89,-0.95 &0.150\\
        \textbf{Texas} &5 &1703 &183 &279 &0.86,-0.86
        &0.097\\
        \textbf{Cornell} &5 &1703 &183 &277 &0.81,-0.93 &0.386\\
        \hline
\end{tabular}
\end{table}

\begin{table*}[t]
\centering
\setlength{\tabcolsep}{2pt} 
\renewcommand{\arraystretch}{1.5} 
\caption{Test accuracy scores(\%) in six heterophily graph benchmarks. The highest accuracies are highlighted in  \textbf{bold}}
\label{heterophily}
\scalebox{0.8}{
\begin{tabular}{lllllll}
\hline
 \textbf{Methods}                            &  \textbf{Cornell}     &  \textbf{Wisconsin}                   &  \textbf{Texas}       &  \textbf{Actor}        &  \textbf{Chameleon}   &  \textbf{Squirrel}    \\ \hline
 {MLP-2}                              &         91.30\(\pm\)0.70              &         {93.87\(\pm\)3.33}             &       \textbf{{92.26\(\pm\)0.71}}                &      38.58\(\pm\)0.25                 &     46.72\(\pm\)0.46                  &          31.28\(\pm\)0.27             \\ 
 {GAT}                               &      76.00\(\pm\)1.01                &    71.01\(\pm\)4.66                                  &    78.87\(\pm\)0.86                  &    35.98\(\pm\)0.23                    &    \textbf{63.90\(\pm\)0.46}                 &      42.72\(\pm\)0.33                 \\
 {APPNP}                              &         91.80\(\pm\)0.63              &       92.00\(\pm\)3.59                                &      91.18\(\pm\)0.70                 &     38.86\(\pm\)0.24                  &        51.91\(\pm\)0.56               &     34.77\(\pm\)0.34                  \\
 {H2GCN}                              &     86.23\(\pm\)4.71                &        87.50\(\pm\)1.77                               &     85.90\(\pm\)3.53                  &    38.85\(\pm\)1.77                   &   52.30\(\pm\)0.48                   &       30.39\(\pm\)1.22                \\
 {GCN}                              &   66.56\(\pm\)13.82                    &       66.72\(\pm\)1.37                                &    75.66\(\pm\)0.96                   &    30.59\(\pm\)0.23                   &    60.96\(\pm\)0.78                   &   45.66\(\pm\)0.39                    \\
 {Mixhp}                              &    60.33\(\pm\)28.53                  &     77.25\(\pm\)7.80                                  &    76.39\(\pm\)7.66                   &      33.13\(\pm\)2.40                 &     36.28\(\pm\)10.22                  &    24.55\(\pm\)2.60                   \\
 {GraphSAGE}                          &      71.41\(\pm\)1.24                 &    64.85\(\pm\)5.14                                    &      79.03\(\pm\)1.20                 &     36.37\(\pm\)0.21                   &    62.15\(\pm\)0.42                   &   41.26\(\pm\)0.26                    \\ 

 {UFG} &    {78.25\(\pm\)0.25} &   {91.01\(\pm\)1.55} &    {82.25\(\pm\)0.91} &    {37.21\(\pm\)0.29} &   56.81\(\pm\)0.12 &   44.28\(\pm\)0.91 \\ \hline

 {RC-UFG (Hom)} &    {77.60\(\pm\)0.81} &   {90.01\(\pm\)1.91} &    {75.75\(\pm\)2.96} &    {35.91\(\pm\)0.88} &   {55.01\(\pm\)1.99} &   {40.59\(\pm\)3.20} \\

 {RC-UFG (Het)} &    {90.31\(\pm\)0.44} &   {91.29\(\pm\)2.81} &    {79.96\(\pm\)1.38} &    {38.91\(\pm\)0.91} &   {60.12\(\pm\)0.74} &   {44.53\(\pm\)1.53} \\ 

 {RC-UFG (Het)\_D} &    \textbf{92.75\(\pm\)0.41} &   \textbf{93.91\(\pm\)2.01} &    {81.05\(\pm\)0.21} &    \textbf{39.41\(\pm\)0.83} &   {61.12\(\pm\)0.25} &   \textbf{46.58\(\pm\)0.44} \\ 
 
 \hline
\end{tabular}}
\end{table*}

\paragraph{Results}
Based on Table \ref{heterophily}, both RC-UFG (Het) and RC-UFG (Het)\_D achieve state-of-the-art performance in most of heterophilic benchmarks. Specifically, we observed that the original framelet (UFG) also delivers a good performance in several heterophily graphs such as \textbf{Wisconsin} and \textbf{Squirrel} indicates HFD nature of framelet convolution. Furthermore, the learning accuracy of RC-UFG (Hom) is in general less than original framelet, suggesting the additional smoothing effect from RC-UFG (Hom) is non-needed in terms of heterophily graph learning. Lastly, the edge drop assisted RC-UFG (Het) achieves the best learning outcomes in most of dataset especially when the number of edge becomes larger, this observation aligns with our claim in Remark \ref{soft_hard} that is: when the graph becomes denser, a stronger sharpening effect (i.e, from CBED) on framelet is needed.

\subsection{Ablation Study}\label{sec:ablation}

In this section, we present an ablation study to evaluate the effectiveness of incorporating graph curvature information into state-of-the-art models, namely GCN \cite{kipf2016semi} and APPNP \cite{DBLP:conf/iclr/KlicperaBG19}. Our previous empirical findings have demonstrated the adaptability of our proposed model by appropriately assigning curvature information to graph framelets, accommodating both homophily and heterophily graphs. We now investigate the impact of integrating graph curvature information on classification accuracy and compare it with the performance of GCN and APPNP. Notably, when assisted by curvature information, both ablation models consistently outperform their baseline results, although to a lesser extent compared to the improvement observed within our proposed model. Additionally, we explore the effectiveness of CBED within the framework of GCN and APPNP. The experiments encompass both homophily datasets (Cora and Pubmed) and heterophily datasets (Texas and Wisconsin), while maintaining consistent hyperparameters and splitting ratios employed in previous sections. For clarity in this section, we report results for heterophily graphs with one decimal point. The results are presented in Table \ref{ablation}.

\begin{table}[t]
\centering
\setlength{\tabcolsep}{2pt} 
\renewcommand{\arraystretch}{1.5} 
\caption{Test accuracy scores(\%) of ablation study by incorporating curvature based information or graph rewiring (edge drop) to two classic baseline models.}
\label{ablation}
\begin{tabular}{cclll}
\hline
\textbf{Methods} & \textbf{Cora}        & \textbf{Pubmed} & \textbf{Texas} & \textbf{Wisconsin} \\ \hline
GCN              & \multicolumn{1}{l}{} 81.5\(\pm\)0.5    &       79.0\(\pm\)0.3       &      75.7\(\pm\)0.9           &   66.7\(\pm\)1.3                 \\
GCN(Hom)         &  82.6\(\pm\)0.4                    &  79.9\(\pm\)0.7               &  72.0\(\pm\)1.3                &   61.3\(\pm\)1.9                 \\
GCN(Het)         & 80.9\(\pm\)0.5                      & 78.3\(\pm\)0.6                 &     77.1\(\pm\)0.5             & 68.4\(\pm\)1.2                   \\
GCN(Het)\_D      & \multicolumn{1}{l}{} 80.0\(\pm\)0.6 & 77.4\(\pm\)0.8                 &    77.8\(\pm\)0.6              &   68.9\(\pm\)0.5                 \\ \hline
APPNP            & \multicolumn{1}{l}{} 83.5\(\pm\)0.7 & 80.2\(\pm\)0.3                 &  91.2\(\pm\)0.7                & 92.0\(\pm\)3.6                   \\
APPNP(Hom)       &  83.9\(\pm\)0.5                    &   80.8\(\pm\)0.2              &   89.3\(\pm\)0.9               & 88.4\(\pm\)1.6                   \\
APPNP(Het)       &  81.8\(\pm\)0.7                    &  79.3\(\pm\)0.7               &   92.1\(\pm\)0.6               &    93.1\(\pm\)2.5                \\
APPNP(Het)\_D    &  81.3\(\pm\)0.8                    &  78.7\(\pm\)0.5               &    92.8\(\pm\)0.7              &  93.6\(\pm\)1.8                  \\ \hline
\end{tabular}
\end{table}

\paragraph{Results}
The results of the ablation study, as presented in Table \ref{ablation}, highlight the impact of assigning curvature information in improving the prediction accuracy of the baseline models. When $\zeta^\text{Hom}$ is assigned, both baseline models demonstrate improved accuracy compared to their initial results without curvature information. Furthermore, the smoothing effect induced from $\zeta^\text{Hom}$ leads to reduced accuracy for the baseline models when the input graph exhibits heterophily. However, the introduction of curvature information based on $\zeta^\text{Het}$ or $\zeta^\text{Hom}$ with CBED, not only restores the models' performance on heterophily graphs but also yields notable increases in accuracy. These observations suggest that curvature-based graph information, or graph rewiring, not only benefits the graph framelet model but also demonstrates its potential in enhancing other GNNs. It is worth noting that while both GCN and APPNP show performance improvements, these increases are comparatively less significant than those observed in the framelet model. This discrepancy is likely attributed to the ability of graph framelets to partially adapt to both types of graph datasets due to energy dynamic reasons \cite{han2022generalized}. However, we acknowledge that a more detailed discussion regarding this matter is beyond the scope of this paper and leave it for future research.

\subsection{Computational Complexity of Graph Ricci Curvature}\label{sec:computation_complexity}

The computation of graph Ricci curvature for large graphs can be computationally expensive due to the requirement of solving a linear programming problem for each edge of the graph. \cite{ye2019curvature} have highlighted that obtaining the Wasserstein distance between probability measure functions on each edge involves a linear programming process with $d_x \times d_y$ variables and $d_x + d_y$ constraints, where $d$ represents the degree of the node. When employing the interior point solver (ECOS) and considering the computational complexity as $\mathcal{O}((d_x \times d_y)^w)$, with $w$ denoting the exponent of the complexity of matrix multiplication (currently known to be 2.373). Fortunately, there are approximation methods available to mitigate this burden, such as the Sinkhorn Algorithm \cite{cuturi2013sinkhorn}. These methods alleviate the main computational cost associated with graph Ricci curvature, making it more feasible to compute for large graphs.


\section{Concluding Remarks and Future Studies}
\label{conclusion}
We explored how the graph geometric information can be utilized to enhance the performance of existing GNNs. The transformed graph Ricci curvature reveals the dynamics between graph local patches, based on which the Ricci Curvature Laplacian can be incorporated into graph convolution and greatly improve the performance. This was proved by both theoretical verification and extensive numeric experiments  where our proposed model outperforms baselines via different types of graph datasets. The positive results show the great potential and encourage us to explore it further. Specifically, one may interested in exploring the methodology of generating an optimized transformation of the graph Ricci curvature to maximize the information availability for downstream learning tasks. In addition, apart from the optimized $\zeta$, another interesting aspect can be consider is to explore the equivalence between curvature information based framelet model (i.e., RC-UFG(Hom) and RC-UFG(Het)) and the original framelet model via energy dynamic perspective. For example, since we have shown by assigning different form of $\zeta$, framelet's energy dynamic will be modified accordingly, if the modified energy dynamic can be approximated by simply utilizing framelet itself, then one could show the curvature enhanced framelet model in fact introduced some restrictions on the learnable matrices (i.e., $\boldsymbol{\theta}$) that controls its dynamic.



\bibliographystyle{plain}
\bibliography{ref}

\end{document}